\documentclass{article}

\PassOptionsToPackage{numbers}{natbib}

\usepackage[final]{neurips_2022}

\usepackage[utf8]{inputenc} 
\usepackage[T1]{fontenc}    
\usepackage{xcolor}         
\usepackage{url}            
\usepackage{hyperref}

\usepackage{booktabs}       
\usepackage{amsfonts}       
\usepackage{nicefrac}       
\usepackage{microtype}      
\usepackage{multirow}

\usepackage{amsmath,amsfonts,bm}

{}
{}
{}
{}

\def\eqref#1{equation~\ref{#1}}

\def\1{\bm{1}}

\def\vf{{\bm{f}}}
\def\vg{{\bm{g}}}
\def\vh{{\bm{h}}}

\def\vk{{\bm{k}}}

\def\vq{{\bm{q}}}

\def\vs{{\bm{s}}}

\def\vu{{\bm{u}}}
\def\vv{{\bm{v}}}

\def\vx{{\bm{x}}}
\def\vy{{\bm{y}}}

\def\mF{{\bm{F}}}

\def\mW{{\bm{W}}}

\DeclareMathAlphabet{\mathsfit}{\encodingdefault}{\sfdefault}{m}{sl}
\SetMathAlphabet{\mathsfit}{bold}{\encodingdefault}{\sfdefault}{bx}{n}
\newcommand{\tens}[1]{\bm{\mathsfit{#1}}}

\def\tF{{\tens{F}}}

\newcommand{\softmax}{\mathrm{softmax}}
\newcommand{\ODESolve}{\mathrm{ODESolve}}

\newcommand{\FFN}{\mathrm{FFN}}
\newcommand{\FWP}{\mathrm{FWP}}

\usepackage{graphicx}
\usepackage{subfig}

\title{Neural\hspace{-0.1mm} Differential\hspace{-0.1mm} Equations for\hspace{-0.3mm} Learning\hspace{-0.3mm} to\hspace{-0.3mm} Program Neural Nets Through Continuous Learning Rules}

\author{
Kazuki Irie$^{1}$ ~ Francesco Faccio$^{1}$ ~ J\"urgen Schmidhuber$^{1,2}$\\
  $^1$The Swiss AI Lab, IDSIA, USI \& SUPSI, Lugano, Switzerland \\
  $^2$AI Initiative, KAUST, Thuwal, Saudi Arabia \\
  \texttt{\{kazuki, francesco, juergen\}@idsia.ch} \\
}

\begin{document}

\maketitle

\begin{abstract}
Neural ordinary differential equations (ODEs) have attracted much attention as
continuous-time counterparts of deep residual neural networks (NNs),
and numerous extensions for recurrent NNs have been proposed.
Since the 1980s,
ODEs have also been used to derive theoretical results for NN learning rules,
e.g., the famous connection between Oja's rule and principal component analysis.
Such rules are typically expressed as additive iterative update processes which have  straightforward ODE counterparts.
Here we introduce a novel combination of learning rules and Neural ODEs to build continuous-time sequence processing nets that learn to manipulate short-term memory in rapidly changing synaptic connections of other nets.
This yields continuous-time counterparts of Fast Weight Programmers and linear Transformers.
Our novel models outperform the best existing Neural Controlled Differential Equation based models on various time series classification tasks,
while also addressing their fundamental scalability limitations.
Our code is public.\footnote{\url{https://github.com/IDSIA/neuraldiffeq-fwp}}
\end{abstract}

\section{Introduction}
\label{sec:intro}
Neural ordinary differential equations (NODEs) \citep{ChenRBD18}
have opened a new perspective on continuous-time computation with neural networks (NNs) as a practical framework for machine learning based on differential equations.
While the original approach---proposed as a continuous-depth version of deep feed-forward residual NNs \citep{resnet, srivastava2015icml}---only covers autonomous ODEs entirely determined by the initial conditions,
more recent 
extensions deal with sequential data (reviewed in Sec.~\ref{sec:back_ode}) in a way similar to what is typically done with standard recurrent NNs (RNNs) in the discrete-time scenario.
This potential for continuous-time (CT) sequence processing (CTSP) is particularly interesting, since there are many applications where datapoints are observed at irregularly spaced time steps,
and CT sequence models might better deal with such data than their discrete-time counterparts.
However, the development of NODEs for CTSP is still at an early stage.
For example, a popular approach of Neural Controlled Differential Equations \citep{KidgerMFL20} (NCDEs; also reviewed in Sec.~\ref{sec:back_ode})
has in practice only one architectural variant corresponding to the ``vanilla'' RNN \citep{elman1990finding}.
Discrete-time processing, however, exploits many different RNN architectures as well as Transformers \citep{trafo}.

While it is not straightforward to transform the standard Transformer into a CT sequence processor,
we'll show that the closely related Fast Weight Programmers (FWPs) \citep{Schmidhuber:91fastweights, schmidhuber1992learning, schlag2021linear, irie2021going} and linear Transformers \citep{katharopoulos2020transformers} (reviewed in Sec.~\ref{sec:back_fwp}) have direct CT counterparts.
In FWPs, temporal processing of short-term memory (stored in fast weight matrices) uses learnable sequences of \textit{learning rules}. Hence
CT versions of FWPs will require differential equations to model the learning rules.
This relates to a trend of the 1980s/90s.
Among many old connections between NNs and dynamical systems described by ODEs (e.g., \citep{funahashi1993approximation, lapedes1987nonlinear, pineda1987generalization,  pearlmutter1989learning, sato1991learning, rico1992discrete}),
the theoretical analysis of NN learning rules in the ODE framework has been particularly fruitful.
Consider the famous example of Oja's rule \citep{oja1982simplified} (briefly reviewed in Sec.~\ref{sec:back_lr}): many results on its stability, convergence, and connection to Principal Component Analysis \citep{pearson1901liii, hotelling1933analysis} were obtained using its ODE counterpart (e.g., \citep{oja1982simplified, oja1985stochastic, oja1989neural, plumbley1995lyapunov, hornik1992convergence, sanger1989optimal, WyattE95, fort1996convergence}).

Here we propose a novel combination of Neural ODEs
and learning rules, to
obtain a new class of sequence processing Neural ODEs which are continuous-time counterparts of Fast Weight Programmers and linear Transformers.
The resulting models are general-purpose CT sequence-processing NNs, which can directly replace the standard Neural CDE models typically used for supervised CT sequence processing tasks.
To the best of our knowledge,
there is no previous work on Neural ODE-based Transformer families for CT sequence processing, despite their popularity in important types of discrete time computation such as Natural Language Processing and beyond.
We also show how our approach solves the fundamental limitation of  existing Neural CDEs in terms of model size scalability.

We conduct experiments on three standard
time series classification tasks
covering various scenarios (regularly sampled, irregularly sampled with missing values, and very long time series).
We demonstrate that our novel models outperform existing Neural ODE-based sequence processors, in some cases by a large margin.

\section{Background}
\label{sec:background}
We briefly review the main background concepts this work builds upon: NODEs for sequence processing (Sec.~\ref{sec:back_ode}), NN learning rules and their connection to ODEs (Sec.~\ref{sec:back_lr}), and Fast Weight Programmers whose memory update is based on learning rules controlled by an NN (Sec.~\ref{sec:back_fwp}).

\subsection{Neural ODEs (NODEs) and Their Extensions for Sequence Processing}
\label{sec:back_ode}

Here we review the core idea of NODEs \citep{ChenRBD18}.
In what follows, let $n$, $N$, $d$, $d_\text{in}$ denote positive integers, $T$ be a positive real number, and $\theta$ denote an arbitrary set of real numbers.
We consider a
residual layer (say, the $n$-th layer with a dimension $d$) in an $N$-layer deep NN 
which transforms an input $\vh_{n-1} \in \mathbb{R}^{d}$ to an output $\vh_{n} \in \mathbb{R}^{d}$ with a parameterised function $\vf_\theta: \mathbb{R}^{d} \rightarrow \mathbb{R}^{d}$ as follows:
\begin{eqnarray}
\label{eq:resnet}
\vh_n = \vh_{n-1} + \vf_\theta(\vh_{n-1})
\end{eqnarray}
This coincides \citep{weinan2017proposal, Haber_2017, HaberRHJ18, ChangMHRBH18, LuZLD18, ChangMHTB18, CicconeGMOG18, ChenRBD18} with the following equation for $\epsilon=1$
\begin{eqnarray}
\vh({t_n}) = \vh({t_{n-1}}) + \epsilon \vf_\theta(\vh({t_{n-1}}))
\end{eqnarray}
where $\vh: [0, T] \rightarrow \mathbb{R}^d$ is a function such that $\vh(t_n) = \vh_n$ holds for all $n: 0\leq n \leq N$ and $t_n \in [0, T]$ such that $t_n-t_{n-1} = \epsilon > 0$ if $n \geq 1$.
This equation is a forward Euler discretisation of the ordinary differential equation defined for all $t \in (t_0, T]$ as
\begin{eqnarray}
\vh'(t) = \vf_\theta(\vh(t)) \quad \text{or} \quad \vh(t) = \vh(t_0) + \int_{s=t_0}^t \vf_\theta(\vh(s))ds
\end{eqnarray}
where $\vh'$ denotes the first order derivative.
This establishes the connection between the ODE and the deep residual net with parameters $\theta$ shared across layers\footnote{Or we make $\theta$ dependent of $t$ such that  parameters are ``depth/layer-dependent'' as in  standard deep nets.}:
given the initial condition $\vh(t_0)=\vh_0$,
the solution to this equation evaluated at time $T$, i.e., $\vh(T)$, corresponds to the output of this deep residual NN, which can be computed by an ODE solver.
We denote it
as a function $\ODESolve$ taking four variables: $\vh(T) = \ODESolve(\vf_\theta, \vh_0, t_0, T)$.
During training, instead of backpropagating through the ODE solver's operations,
the continuous \textit{adjoint sensitivity method} \citep{pontryagin1962}
(which essentially solves another ODE but backward in time) can compute gradients with $O(d)$ memory requirement, constant w.r.t.~$T$ \citep{ChenRBD18}.

A natural next step is to extend this formulation for RNNs,
i.e., the index $n$ now denotes the time step,
and we assume an external input $\vx_n \in \mathbb{R}^{d_\text{in}}$ at each step $n$ to update the hidden state $\vh_{n-1}$ to $\vh_n$ as
\begin{eqnarray}
\label{eq:rnn}
\vh_n = \vf_\theta(\vh_{n-1}, \vx_n)
\end{eqnarray}

Depending on the property of external inputs $(\vx_n)_{n=1}^N = (\vx_1,..., \vx_N)$, there are different ways of defining NODEs for sequence processing.
We mainly distinguish three cases.

\textbf{First}, when there is a possibility to construct a \textit{differentiable} control signal $\vx:t\mapsto \vx(t) \in \mathbb{R}^{d_\text{in}}$ for $t \in [t_0, T]$ from the inputs $(\vx_n)_{n=1}^N$; an attractive approach by
\citet{KidgerMFL20} handles the corresponding dynamics in a \textit{neural controlled differential equation} (NCDE):
\begin{eqnarray}
\label{eq:cde}
\vh(t) = \vh(t_0) + \int_{s=t_0}^t \mF_\theta(\vh(s))d\vx(s) = \vh(t_0) + \int_{s=t_0}^t \mF_\theta(\vh(s))\vx'(s)ds
\end{eqnarray}
where $\mF_\theta$ is a parameterised function (typically a few-layer NN) which maps a vector $\vh(s) \in \mathbb{R}^d$ to a matrix $\mF_\theta(\vh(s)) \in  \mathbb{R}^{d \times d_{\text{in}}}$ (we already relate this component to Recurrent Fast Weight Programmers below) and thus, $\mF_\theta(\vh(s))d\vx(s)$ denotes a matrix-vector multiplication.
There are several methods to construct the \textit{control} $\vx: [t_0, T] \rightarrow \mathbb{R}^{d_\text{in}}$ based on the discrete data points $(\vx_n)_{n=1}^N$, such that its differentiability is guaranteed.
In this work, we follow \citet{KidgerMFL20} and mainly use natural cubic splines over all data points
(which, however, makes it incompatible with auto-regressive processing); for better alternatives, we refer to \citet{morrill2021neural}.
Since the final equation is again an NODE with a vector field of form $\vg_{\theta, \vx'}(s, \vh(s)) = \mF_\theta(\vh(s))\vx'(s)$,
all methods described above are applicable: ODE solver for evaluation and continuous adjoint method for memory efficient training.
A notable extension
of Neural CDEs is the use of \textit{log-signatures} to sub-sample the input sequence \citep{morrillSKF21}.
The resulting NCDEs are called \textit{neural rough differential equations} (NRDEs),
which are relevant for processing long sequences.
One fundamental limitation of the NCDEs above is the lack of scalability of $\mF_\theta: \mathbb{R}^{d} \rightarrow \mathbb{R}^{d \times d_{\text{in}}}$.
For example, if we naively parameterise $\mF_\theta$ using a linear layer, the size of its weight matrix is $d^2*d_{\text{in}}$ which quadratically increases with the hidden state size $d$.
Previous attempts \citep{kidger2022neural} do not successfully resolve this issue without performance degradation.
In Sec.~\ref{sec:diss}, we'll discuss how our models (Sec.~\ref{sec:cde-lr}) naturally circumvent this limitation while remaining powerful NCDEs.

On a side note, the NCDE is often referred to as the ``continuous-time analogue'' to RNNs \citep{KidgerMFL20}, but this is a bit misleading:
discrete-time RNN equations corresponding to the continuous-time Eq.~\ref{eq:cde}
do not reflect the standard RNN of Eq.~\ref{eq:rnn}
but:
\begin{align}
\label{eq:cde_discrete_1}
\vh_n &= \vh_{n-1} + \mW_{n-1} (\vx_n -\vx_{n-1}) \\
\label{eq:cde_discrete_2}
\mW_{n} & =  \mF_\theta(\vh_{n})
\end{align}
where one network (Eq.~\ref{eq:cde_discrete_1}) learns to translate the variation of inputs $(\vx_n -\vx_{n-1})$ into a change in the state space, using a weight matrix $\mW_{n-1}$ which itself is generated by another network ($\mF_\theta : \mathbb{R}^{d} \rightarrow \mathbb{R}^{d \times d_\text{in}}$; Eq.~\ref{eq:cde_discrete_2}) on the fly from the hidden state.
This model is thus a kind of Recurrent FWP \citep{schmidhuber1993reducing, ha2017hypernetworks, irie2021going}.

\textbf{Second}, even if $\vx$ is not differentiable, having access to (piece-wise) continuous $\vx$ defined and bounded over an interval of interest $[t_0, T]$ is enough to define a sequence processing NODE, by making it part of the vector field:
\begin{eqnarray}
\label{eq:direct-ode}
\vh(t) = \vh(t_0) + \int_{s=t_0}^t \vf_\theta(\vh(s), \vx(s))ds
\end{eqnarray}
where the vector field $\vf_\theta(\vh(t), \vx(t)) = \vg_{\theta, \vx}(t, \vh(t))$
can effectively be evaluated at any time $t \in [t_0, T]$.
We refer to this second approach as a \textit{direct NODE} method.
While \citet{KidgerMFL20} theoretically and empirically show that this approach is less expressive than the NCDEs above,
we'll show how in our case of learning rules one can derive interesting models within this framework, which empirically perform on par with the CDE variants.

\textbf{Finally}, when no control function with one of the above properties can be constructed,
a mainstream approach dissociates
the continuous-time hidden state update via ODE for the time between two observations (e.g., Eq.~\ref{eq:ode-rnn-ode} below) from
integration of the new data (Eq.~\ref{eq:ode-rnn-rnn} below).
Notable examples of this category include ODE-RNNs \citep{RubanovaCD19, BrouwerSAM19}
which transform the hidden states $\vh_{n-1}$ to $\vh_n$ for each observation $\vx_n$ available at time $t_n$ as follows:
\begin{eqnarray}
\label{eq:ode-rnn-ode}
\vu_n &=& \ODESolve(\vf_{\theta_1}, \vh_{n-1}, t_{n-1}, t_n) \\
\label{eq:ode-rnn-rnn}
\vh_n &=& \bm{\phi}_{\theta_2}(\vx_{n}, \vu_{n})
\end{eqnarray}
where Eq.~\ref{eq:ode-rnn-ode} autonomously updates the hidden state between two observations using a function $\vf_{\theta_1}$ parameterised by $\theta_1$, while
in Eq.~\ref{eq:ode-rnn-rnn},
function $\bm{\phi}_{\theta_2}$ parameterised by $\theta_2$ 
integrates the new input $\vx_n$ into the hidden state.
In Latent ODE-RNN \citep{RubanovaCD19}, a popular extension of this approach to the variational setting,  the initial recurrent state $\vh_0$
is sampled from a prior (during training, an additional encoder is trained to map sequences of inputs to parameters of the prior).
While this third case is not our focus, we'll also show how to use FWPs in this scenario in Sec.~\ref{sec:ode_rfwp} for the sake of completeness.

\subsection{Learning Rules and Their Connections to ODEs}
\label{sec:back_lr}
Learning rules of artificial NNs describe the process which modifies their weights in response to some inputs.
This includes the standard backpropagation rule (also known as the reverse mode of automatic differentiation) derived for the case of supervised learning, as well as rules inspired by Hebb's informal rule \citep{hebb1949organization} in ``unsupervised'' settings.
Here we focus on the latter.
Let $n$, $d_\text{in}$ $d_\text{out}$ be positive integers.
Given a linear layer with a weight matrix $\mW_{n} \in \mathbb{R}^{d_\text{out} \times d_\text{in}}$ (the single output neuron case $d_\text{out}=1$ is the focus of the classic works) at time $n$ which transforms input $\vx_{n} \in \mathbb{R}^{d_\text{in}}$ to output $\vy_{n} \in \mathbb{R}^{d_\text{out}}$ as
\begin{eqnarray}
\label{eq:hebb}
\vy_{n} &=& \mW_{n-1} \vx_n
\end{eqnarray}
the pure Hebb-style additive learning rule modifies the weights according to
\begin{eqnarray}
\label{eq:hebb_update}
\mW_n &=& \mW_{n-1} + \eta_n \vy_{n} \otimes \vx_n
\end{eqnarray}
where $\otimes$ denotes outer product and $\eta_n \in \mathbb{R}_{+}$ is a learning rate at time $n$.

\citet{oja1982simplified} proposed stability improvements to this rule through a decay term
\begin{eqnarray}
\label{eq:oja_82}
\mW_n = \mW_{n-1} + \eta_n \vy_n \otimes (\vx_n - \mW_{n-1}^\top \vy_n)
\end{eqnarray}
whose theoretical analysis has since the 1980s been a subject of many researchers covering stability, convergence, and relation to Principal Component Analysis \citep{oja1982simplified, oja1985stochastic, oja1989neural, plumbley1995lyapunov, hornik1992convergence, sanger1989optimal, WyattE95, fort1996convergence, ChouW20}.
One key approach for such theoretical analysis is to view the equation above as a discretisation of the following ODE:
\begin{eqnarray}
\label{eq:oja_ode_original}
\mW'(t) = \eta(t) \vy(t) \otimes (\vx(t) - \mW(t-1)^\top \vy(t)) 
\end{eqnarray}
On a related note, studies of RNNs (e.g., \citep{zhang2014comprehensive, fermanian2021framing}) or learning dynamics (e.g., \citep{HeuselRUNH17}) have also profited from ODEs.

\subsection{Fast Weight Programmers \& Linear Transformers}
\label{sec:back_fwp}
Fast Weight Programmers (FWP; \citep{Schmidhuber:91fastweights, schmidhuber1992learning, schlag2021linear, irie2021going}) are general-purpose (auto-regressive) sequence processing NNs.
In general,
an FWP is a system of two NNs: a \textit{slow} NN, the \textit{programmer}, rapidly generates during runtime weight changes of another neural network, the \textit{fast} NN.
The (slow) weights of the slow net are typically trained by gradient descent.
Variants of FWPs whose weight generation is based on outer products between keys and values \citep{Schmidhuber:91fastweights} have been shown \citep{schlag2021linear} to be equivalent to Linear Transformers \citep{katharopoulos2020transformers}
(using the mathematical equivalence known from perceptron/kernel machine duality \citep{aizerman1964theoretical, irie2022dual}).
These FWPs use sequences of learning rules to update short-term memory in  form of a fast weight matrix.
A practical example of such FWPs is the DeltaNet \citep{schlag2021linear} which transforms an input $\vx_n \in \mathbb{R}^{d_\text{in}}$ into an output $\vy_n \in \mathbb{R}^{d_\text{out}}$ at each time step $n$ while updating its fast weight matrix $\mW_{n-1} \in \mathbb{R}^{d_\text{out} \times d_\text{key} }$ as follows:
\begin{eqnarray}
\label{eq:proj}
\beta_n, \vq_n, \vk_n, \vv_n &=& \mW_\text{slow}\vx_n\\
\label{eq:lr}
\mW_n &=& \mW_{n-1} + \sigma(\beta_n)(\vv_n - \mW_{n-1} \phi(\vk_n)) \otimes \phi(\vk_n) \\
\vy_n  &=& \mW_n \phi(\vq_n) \label{eq:fw_get}
\end{eqnarray}
where the slow net (Eq.~\ref{eq:proj}; with weights $\mW_\text{slow} \in \mathbb{R}^{(1 + 2 *d_\text{key} + d_\text{out}) \times d_\text{in}}$) generates key/value vectors $\vk_n \in \mathbb{R}^{d_\text{key}}$ and $\vv_n \in \mathbb{R}^{d_\text{out}}$ as well as a scalar $\beta_n \in \mathbb{R}$ to obtain a dynamic learning rate by applying a sigmoid function $\sigma$, and $\phi$ is an element-wise activation function
whose output elements are positive and sum up to one (typically softmax).
These fast dynamic variables generated by a slow NN are used in
a learning rule (Eq.~\ref{eq:lr}) akin to the classic delta rule \citep{widrow1960adaptive}
to update the fast weight matrix.
The output is finally produced by the forward computation of the fast NN, i.e., by
\textit{querying} the fast weight matrix by the generated query vector $\vq_n \in \mathbb{R}^{d_\text{key}}$ (Eq.~\ref{eq:fw_get}).
An intuitive interpretation of the fast weight matrix is a key-value associative memory with
write and read operations defined by Eq.~\ref{eq:lr} and \ref{eq:fw_get}, respectively.
This encourages intuitive thoughts about memory capacity (limited by the number of ``keys'' we can store without interference) \citep{schlag2021linear}.
For instance, if we replace the learning rule (i.e., memory writing operation) of Eq.~\ref{eq:lr} by a pure additive Hebb-style rule (and a fixed learning rate of 1.0): $\mW_n = \mW_{n-1} + \vv_n \otimes \phi(\vk_n)$, we obtain the Linear Transformer \citep{katharopoulos2020transformers} (we refer to prior work \citep{schlag2021linear} for further explanations of the omission of attention normalisation).
Such a purely additive learning rule often suffers from long term dependencies, unlike the delta rule \citep{schlag2021linear}.
We'll confirm this trend also in the CT models (using the EigenWorms dataset).
For later convenience, we introduce a notation $\FWP$ which denotes generic FWP operations:
$\vy_n, \mW_n = \FWP(\vx_n, \mW_{n-1}; \mW_\text{slow})$.

\section{Continuous-Time Fast Weight Programmers}
\label{sec:ct_fwp}
We propose continuous-time counterparts of Fast Weight Programmers (Sec.~\ref{sec:back_fwp})
which naturally combine ODEs for learning rules (Sec.~\ref{sec:back_lr}) and existing approaches for sequence processing with NODEs (Sec.~\ref{sec:back_ode}).
We present three types of these CT FWP models in line with
the categorisation of Sec.~\ref{sec:back_ode} while the main focus of this work is on the two first cases.

\subsection{Direct NODE-based FWPs}
\label{sec:sub_direct_ode}
In the \textit{direct NODE} approach (reviewed in Sec.~\ref{sec:back_ode}),
we assume a (piece-wise) continuous control signal $\vx: t\mapsto \vx(t)$ bounded over an interval $[t_0, T]$.
We make it part of the vector field to define an ODE describing a \textit{continuous-time learning rule} for a fast weight matrix $\mW(t)$:
\begin{eqnarray}
\label{eq:ode-lr}
\mW(t) = \mW(t_0) + \int_{s=t_0}^t \mF_\theta(\mW(s), \vx(s))ds
\end{eqnarray}
where $\mW: t\mapsto \mW(t) \in \mathbb{R}^{d_\text{out} \times d_\text{key}}$ is a function defined on $[t_0, T]$,
and $\mF_\theta$ is an NN parameterised by $\theta$ which maps onto $\mathbb{R}^{d_\text{out} \times d_\text{key}}$.
This is a neural differential equation for 
learning to program a neural net through continuous learning rules, that is,
to train a fast weight matrix $\mW(t)$ of a fast NN (Eq.~\ref{eq:querying_2} below) for each sequential control $\vx$. 
Like in the discrete-time FWPs (Sec.~\ref{sec:back_fwp}), the output $\vy(T) \in \mathbb{R}^{d_\text{out}}$ is obtained by \textit{querying} this fast weight matrix\footnote{In practice, we also apply element-wise activation functions to query/key/value vectors where appropriate, which we omit here for readability. We refer to Appendix \ref{app:model} for further details.} (e.g., at the last time step $T$):
\begin{align}
\label{eq:querying_1}
\vq(T) &= \mW_q \vx(T) \\
\label{eq:querying_2}
\vy(T) &= \mW(T) \vq(T)
\end{align}
where $\mW_q \in \mathbb{R}^{d_\text{key} \times d_\text{in}}$ is a slow weight matrix used to generate the query $\vq(T) \in \mathbb{R}^{d_\text{key}}$ (Eq.~\ref{eq:querying_1}).
Now we need to specify $\mF_\theta$ in Eq.~\ref{eq:ode-lr} to fully define the learning rule.
We focus on three variants:
\begin{align} 
\label{eq:direct_ode}
\mF_\theta(\mW(s), \vx(s))
    &= \sigma(\beta(s))
    \begin{cases}
      \vk(s) \otimes \vv(s) & \text{Hebb-style}\\
      \vv(s) \otimes \big(\vk(s) - \mW(s)^\top \vv(s) \big) & \text{Oja-style}\\
      \big(\vv(s) - \mW(s)\vk(s)\big) \otimes \vk(s)& \text{Delta-style}
    \end{cases}
\end{align}
where $[\beta(s), \vk(s), \vv(s)] = \mW_\text{slow} \vx(s)$ with a slow weight matrix $\mW_\text{slow} \in \mathbb{R}^{(1 + d_\text{key} + d_\text{out}) \times d_\text{in}}$.
As in the discrete-time FWP (Sec.~\ref{sec:back_fwp}), the slow NN generates
$\beta(s) \in \mathbb{R}$ (to which we apply the sigmoid function $\sigma$ to obtain a learning rate),
key $\vk(s) \in \mathbb{R}^{d_\text{key}}$ and value $\vv(s) \in \mathbb{R}^{d_\text{out}}$ vectors from input $\vx(s)$.
These variants are inspired by the respective classic learning rules of the same name, while they are crucially different from the classic ones in the sense that all variables involved (key, value, learning rate)
are continually generated by the slow NN.
In the experimental section, we'll comment on how some of these design choices can result in task-dependent performance gaps.
In practice, we use the \textit{multi-head} version of the operations above (i.e., by letting $H$ be a positive integer denoting the number of heads, query/key/value vectors are split into $H$ sub-vectors and Eqs.~\ref{eq:querying_2}-\ref{eq:direct_ode} are conducted independently for each head). The output is followed by the standard feed-forward block like in Transformers \citep{trafo}.
Possible extensions for deeper models are discussed in Appendix \ref{app:deeper}.

\subsection{NCDE-based FWPs}
\label{sec:cde-lr}
Here we present models based on NCDEs (reviewed in Sec.~\ref{sec:back_ode}).
We assume availability of a differentiable control signal $\vx(t)$, whose first order derivative is denoted by $\vx'(t)$.
Given the NCDE formulation of Eq.~\ref{eq:cde},
the most straight-forward approach to obtain a CT Fast Weight Programmer is to extend the dimensionality of the recurrent hidden state, i.e., we introduce
a parameterised function $\tF_\theta$ which maps a matrix $\mW(t) \in \mathbb{R}^{d_\text{out} \times d_\text{key}}$
to a third-order tensor  $\tF_\theta(\mW(t)) \in \mathbb{R}^{d_\text{out} \times d_\text{key} \times d_\text{in}}$:
\begin{eqnarray}
\label{eq:cde-lr}
\mW(t) = \mW(t_0) + \int_{s=t_0}^t \tF_\theta(\mW(s))d\vx(s) = \mW(t_0) + \int_{s=t_0}^t \tF_\theta(\mW(s))\vx'(s)ds
\end{eqnarray}
However, this approach is obviously not scalable since the input and output dimensions ($d_\text{out} \times d_\text{key}$ and $d_\text{out} \times d_\text{key} \times d_\text{in}$) of $\tF_\theta$ can be too large in practice.
A more tractable CDE-based approach can be obtained by providing $\vx$ and/or $\vx'$ to the vector field:
\begin{eqnarray}
\label{eq:cde-lr-x-dx}
\mW(t) = \mW(t_0) + \int_{s=t_0}^t \tF_\theta(\mW(s), \vx(s), \vx'(s))\vx'(s)ds
\end{eqnarray}
While this equation still remains a CDE because of the multiplication from the right by $d\vx=\vx'(s)ds$,
the additional inputs to the vector field offer a way of making use of various learning rules, as in the case of direct NODE approach above (Sec.~\ref{sec:sub_direct_ode}).
To be specific, either $\vx$ and $\vx'$ or only $\vx'$ is required in the vector field to obtain these tractable CDEs.
Here we present the version which uses both $\vx$ and $\vx'$ \footnote{\label{foot:var}
The equations for the version using only $\vx'$ can be obtained by replacing $\vx$ by $\vx'$ in Eq.~\ref{eq:cde_lr}.
We provide an ablation in Appendix \ref{app:cde_variants}.
As a side note, we also obtain the equation for the CDE using only $\vx'$ by replacing $\vx$ by $\vx'$ in Eq.~\ref{eq:ode-lr} for the direct NODE case.\label{foot:node}}.
The resulting vector fields for different cases are:
\begin{align}
\label{eq:cde_lr}
\tF_\theta\big(\mW(s), \vx(s), \vx'(s)\big)\vx'(s)
    &= \sigma(\beta(s))
    \begin{cases}
     \mW_k \vx(s) \otimes \mW_v \vx'(s)  & \text{Hebb}\\
      \big(\mW_k \vx(s) - \mW(s)^\top \mW_v \vx'(s) \big) \otimes \mW_v \vx'(s) & \text{Oja} \\
      \big(\mW_v \vx(s) - \mW(s)\mW_k \vx'(s)\big) \otimes \mW_k \vx'(s) & \text{Delta}
    \end{cases}
\end{align}
As can be seen above, the use of CDEs to describe a continuous fast weight learning rule thus naturally results in a key/value memory where $\vx'$ is used to generate either key or value vectors.
Because of the multiplication from the right by $\vx'$, the role of $\vx'$ changes
depending on the choice of learning rule: $\vx'$ is used to generate the key in the Delta case but the value vector in the case of Oja.
In the case of Hebb, the choice made in Eq.~\ref{eq:cde_lr} of using $\vx$ for keys and $\vx'$ for values is arbitrary since Eq.~\ref{eq:cde_lr} is symmetric in terms of roles of keys and values (see an ablation study in Appendix \ref{app:cde_variants} for the other case where we use $\vx'$ to generate the key and $\vx$ for the value).
The querying operation (analogous to Eqs.~\ref{eq:querying_1}-\ref{eq:querying_2} for the direct NODE case) is also modified accordingly, depending on the choice of learning rule, such that the same input ($\vx$ or $\vx'$) is used to generate both  key and query:
\begin{align} 
\label{eq:cde_fwp_query}
\vy(T) &=
    \begin{cases}
     \mW(T)^\top \mW_q \vx(T)& \text{Hebb and Oja}\\
     \mW(T) \mW_q \vx'(T)  & \text{Delta}
    \end{cases}
\end{align}
Note that since the proposed vector field $\tF_\theta\big(\mW(s), \vx(s), \vx'(s)\big)\vx'(s)$ is more general than the one used in the original NCDE $\tF_\theta\big(\mW(s)\big)\vx'(s)$,
any theoretical results on the CDE remain valid (which, however, does not tell us anything about the best choice for its exact parameterisation).

\subsection{ODE-RFWP and Latent ODE-RFWP}
\label{sec:ode_rfwp}
The main focus of this work is the setting of \citet{KidgerMFL20} where we assume the existence of some control signal $\vx$ (Sec.~\ref{sec:sub_direct_ode} and \ref{sec:cde-lr} above).
However, here we also show a way of using FWPs in the third/last case presented in Sec.~\ref{sec:back_ode} where
no control $\vx(t)$ is available (or can be constructed), i.e., we only have access to discrete observations $(\vx_n)_{n=0}^N$.
Here we cannot directly define the vector field involving continuous transformations using the inputs.
We follow the existing approaches (ODE-RNN or Latent ODE; Sec.~\ref{sec:back_ode}) which use two separate update functions:
A discrete recurrent state update is executed every time a new observation is available to the model, while a continuous update using an autonomous ODE is conducted in between observations.
Unlike with standard recurrent state vectors, however, it is not practical to autonomously evolve high-dimensional fast weight matrices\footnote{
Such an approach would require a computationally expensive matrix-to-matrix transforming NN, whose scalability is limited.
}.
We therefore opt for using a Recurrent FWP (RFWP) \citep{irie2021going} and combine it with an ODE:
\begin{eqnarray}
\label{eq:ode-rfwp-ode}
\vu_n &=& \ODESolve(\vf_{\theta_1}, \vh_{n-1}, t_{n-1}, t_n) \\
\label{eq:ode-rfwp-rnn}
\vh_n, \mW_{n} &=& \FWP([\vx_{n}, \vu_{n}], \mW_{n-1}; \theta_2)
\end{eqnarray}
where we keep the fast weight learning rule itself discrete (Eq.~\ref{eq:ode-rfwp-rnn}), but evolve the recurrent state vector $\vu_n$ using an ODE (Eq.~\ref{eq:ode-rfwp-ode}) such that the information to be read/written to the fast weight matrix is controlled by a variable which is continuously updated between observations.
We refer to this model as ODE-RFWP and its variational variant as Latent ODE-RFWP.

Since we focus on continuous-time learning rules, the case above is not of central interest
as the learning rule remains discrete here (Eq.~\ref{eq:ode-rfwp-rnn}).
Nevertheless, in Appendix \ref{app:rl}, we also provide some experimental results for model-based reinforcement learning settings corresponding to this case.

\section{Experiments}
\label{sec:exp}
We consider three datasets covering three types of time series which are
regularly sampled (Speech Commands \citep{warden2018speech}),
irregularly sampled with partially missing features (PhysioNet Sepsis \citep{ReynaJSJSWSNC19}),
or very long (EigenWorms \citep{bagnall2018uea}).
We compare the proposed direct NODE and CDE based FWP models (Sec.~\ref{sec:sub_direct_ode} \& \ref{sec:cde-lr}) to NODE baselines
previously reported on the same datasets \citep{KidgerMFL20, morrill2021neural}.
Appendix \ref{app:exp_detail} provides further experimental details including hyper-parameters.

\begin{table}[t]
\caption{\textbf{Accuracy (\%) on the Speech Commands classification task} and \textbf{AUC ($\times 10^2$) on the PhysioNet Sepsis prediction task}.
PhysioNet has two cases: with (OI) or without (no-OI) observational intensity (see text for details).
Numbers marked by * are taken from \citet{KidgerMFL20}.
Mean and standard deviation (std) are computed over 5 runs.
}
\label{tab:speech}
\begin{center}
\begin{tabular}{llcrr}
\toprule
\multirow{2}{*}{Type} & \multirow{2}{*}{Model} & \multicolumn{1}{c}{\multirow{2}{*}{Speech Commands}} & \multicolumn{2}{c}{PhysioNet Sepsis} \\ \cmidrule(r){4-5}
& &  & \multicolumn{1}{c}{OI} & \multicolumn{1}{c}{no-OI} \\ \midrule
\multirow{1}{*}{Direct NODE} 
          & GRU-ODE \citep{KidgerMFL20}* & 47.9 (2.9) & 85.2 (1.0) &  77.1 (2.4)   \\ \cmidrule(r){2-5}
          & Hebb    & 82.8 (1.1) & \textbf{90.4 (0.4)} & 82.9 (0.7) \\ 
          & Oja     & \textbf{85.4 (0.9)} & 88.9 (1.4)  & 82.9 (0.5) \\ 
          & Delta   & 81.5 (3.8) & 89.8 (1.0) & \textbf{84.5 (2.9)} \\  \midrule
\multirow{1}{*}{CDE} & NCDE \citep{KidgerMFL20}* &  89.8 (2.5) & 88.0 (0.6) & 77.6 (0.9) \\ \cmidrule(r){2-5}
          & Hebb  & 89.5 (0.3) & 89.9 (0.6) & \textbf{85.7 (0.3)}\\ 
          & Oja   & 90.0 (0.7) & \textbf{91.2 (0.4)} & 85.1 (2.5) \\
          & Delta & \textbf{90.2 (0.2)} & 90.9 (0.2) & 84.5 (0.7) \\
\bottomrule
\end{tabular}
\end{center}
\end{table}

\paragraph{Speech Commands.}
The Speech Commands \citep{warden2018speech} is a single word speech recognition task.
The datapoints are regularly sampled, and the sequence lengths are relatively short ($\leq$ 160 frames), which makes this task a popular sanity check.
Following prior work on NCDEs \citep{KidgerMFL20},
we use 20 mel frequency cepstral coefficients as speech features and
classify the resulting sequence to one out of ten keywords.
The middle column of Table \ref{tab:speech} shows the  results.
The table is split into the direct NODE (top) and CDE (bottom) based approaches.
We first observe that among the direct NODE approaches, all our FWPs largely outperform ($\geq$ 80\% accuracy) the baseline GRU-ODE performance of 47.9\% (the best direct NODE baseline from \citet{KidgerMFL20}).
This demonstrates that with a good parameterisation of the vector field, the direct NODE approach can achieve competitive performance.
On the other hand, all CDE-based approaches yield similar performance.
We also only see slight differences in terms of performance among different learning rules, without a clear winner for this task.
This may indicate that
the ordinary nature of this task (regularly sampled; short sequences) does not allow for differentiating among these CDE models, including the baseline.

\paragraph{PhysioNet Sepsis.}
The PhysioNet Sepsis is a dataset of the sepsis prediction task from the PhysioNet challenge 2019 \citep{ReynaJSJSWSNC19}.
This is again a dataset used by \citet{KidgerMFL20} to evaluate NCDEs.
The task is a binary prediction of sepsis from a time series consisting of measurements of 34 medical features (e.g., respiration rate) of patients' stays at an ICU.
Each sequence is additionally labelled by five static features of the patient (e.g., age)
which are fed to the model to generate the initial state of the ODE.
Sequences are relatively short ($\leq$ 72 frames) but datapoints are irregularly sampled and many entries are missing,
which makes this task challenging.
It comes in two versions: with and without the so-called \textit{observation intensity} information (denoted as ``OI'' and ``no-OI'') which is one extra input feature indicating each observation's time stamp (providing the models with information on measurement frequency).
This distinction is important since the prior work \citep{KidgerMFL20} has reported that existing ODE/CDE-based approaches struggle with the no-OI case of this task.
Following the previous work, we report the performance in terms of Area Under the ROC Curve (AUC).
The right part of Table \ref{tab:speech} shows the results.
We obtain large improvements in the no-IO case (from 77.6 to 85.7\% for the CDEs and from 77.1 to 84.5\% for the direct NODEs),
while also obtaining 
small improvements in the OI case (from 85.2 to 90.4\% for direct NODEs, and from 88.0 to 91.2\% for CDEs).
The no-OI performance of our models is also comparable to the best overall performance reported by \citet{kidger2022neural}: 85.0 \% (1.3) achieved by GRU-D \citep{che2018recurrent}.
This demonstrates the efficacy of
CT FWP model variants for handling irregularly sampled data with partially missing features even in the case without frequency information.
Differences between various learning rules are rather small again.
In some cases, we observe performance to be very sensitive to hyper-parameters. For example, the best Oja-CDE configuration achieves 85.1\% (2.5) with a learning rate of 6e-5, while this goes down to 79.6\% (4.7) when the learning rate is changed to 5e-5.

\begin{table}[t]
\caption{\textbf{Classification Accuracy (\%) on the EigenWorms task.}
Numbers marked by * are taken from \citet{morrillSKF21}.
Mean and standard deviation (std) are computed over 5 runs.
``Sig-Depth'' indicates the depth of the signature (with this number equal to 1, an RDE is reduced to a CDE).
To facilitate comparisons to prior work \citep{morrillSKF21},
we also add the column ``Step'' indicating the sequence down-sampling factor (even if we fix it to the best value \citep{morrillSKF21} of 4).}
\label{tab:eigenworms}
\begin{center}
\begin{tabular}{lccr}
\toprule
Model & Sig-Depth & Step & \multicolumn{1}{c}{Test Acc.~[\%]} \\ \midrule
 NRDE \citep{morrillSKF21}* & 2 & 4 &  83.8 $\;$ (3.0) \\  \midrule 
           Hebb & 2 & 4 & 45.6 $\;$ (5.9) \\ 
           Oja & & &  46.7 $\;$ (7.5) \\
           Delta & &  & \textbf{87.7 $\;$ (1.9)} \\
      \midrule \midrule       
 NCDE \citep{morrillSKF21}*  & 1 & 4 & 66.7 (11.8) \\ \midrule
             Hebb & 1 & 4 &  41.0 $\;$ (6.5)  \\ 
           Oja & & & 49.7 $\;$ (9.9) \\
           Delta & &  & \textbf{91.8} $\;$ (3.4)\\
\bottomrule
\end{tabular}
\end{center}
\end{table}

\paragraph{EigenWorms.}
The EigenWorms dataset (which is part of the UEA benchmark \citep{bagnall2018uea}) is a 5-way classification of roundworm types based on time series tracking their movements.
To be more specific, motions of a worm are represented by six features corresponding to their projections to six template movement shapes, called ``eigenworms.''
While this dataset contains only 259 examples, it is notable for its very long sequences (raw sequence lengths exceed 17\,K)
and long-span temporal dependencies \citep{morrillSKF21, rusch2021unicornn, rusch2021long}.
We use the same train/validation/test split ratio as the prior work \citep{morrillSKF21} which reports Neural RDEs (NRDEs) as achieving the best NODE model performance on this dataset.
The equations of our CT FWPs for the RDE case can be straightforwardly obtained by replacing the input $\vx$ in Eqs.~\ref{eq:ode-lr}-\ref{eq:querying_1} of the direct NODE formulation (or\footnote{Conceptually these two approaches are equivalent: the direct NODE and NCDE coincide here.
In practice, there can be a subtle difference due to an implementation detail.
The direct NODE approach can apply layer normalisation to the input fed to \textit{both} key and value projections (as they are both inside the vector field).
In this case (as in our implementation), the corresponding NCDE formulation we obtain is based on the normalised input.
}, $\vx$ and $\vx'$ in Eqs.~\ref{eq:cde-lr-x-dx}-\ref{eq:cde_fwp_query} of the NCDEs) by the corresponding log-signatures.
Table \ref{tab:eigenworms} shows the results, where
``Step'' denotes the time sub-sampling rate which is fixed to 4 for which the prior work \citep{morrillSKF21} reports the best NRDE and NCDE performance.
``Sig-Depth'' denotes the depth of the log-signature (the deeper, the more log-signature terms we take into account, thus ending up with a larger input feature vector; we refer to the original paper \citep{morrillSKF21} for further details).
We consider two values for this parameter: 1 and 2.
When set to 1, the input feature contains only the first derivative $\vx'(s)$ and thus the NRDE is reduced to an NCDE (with controls constructed via linear interpolation).
We take the best NCDE performance from
\citet{morrillSKF21} as the depth-1 baseline.
\citet{morrillSKF21} report the best overall performance for the depth-2 NRDE (depth-2 baseline in our table).
In both cases, we first note a large performance gap between models with different learning rules.
While the naive Hebb and Oja based models struggle with this very long sequence processing (sequence length still exceeds 4\,K with a down-sampling step size of 4), the Delta rule performs very well.
This confirms the prior result in the discrete-time domain \citep{schlag2021linear}
which motivated the Delta rule design by its potential for 
handling long sequences
(we refer to prior work \citep{schlag2021linear} for further explanations).
Since its performance on other tasks is comparable to the one of Hebb and Oja variants, the Delta rule is a natural default choice for parameterising the CT FWPs.

In both the depth-1 and depth-2 cases, we obtain large improvements compared to the respective baselines.
It is counter-intuitive to find 
certain depth-2 models underperforming their depth-1 counterparts, but this trend has also been observed in the original NRDEs \citep{morrill2021neural}.
Our best overall performance is obtained in the depth-1 case: 91.8 \% (3.4)  exceeds the previous best NRDE based model performance of 83.8 \% (3.0) \citep{morrill2021neural}.
This almost matches the state-of-the-art accuracy of 92.8 \% (1.8) reported  by \citet{rusch2021long} (using an ODE-inspired discrete-time model).
Our model's performance variance is high  (best single seed performance is 97.4\% while the standard deviation is 3.4).
The wall clock time
is similar for our best model (last row in Table \ref{tab:eigenworms}) and the NRDE baseline (36s/epoch on a GeForce RTX 2080 Ti) and their sizes are comparable (87\,K vs.~65\,K parameters respectively).\looseness=-1

\section{Discussions}
\label{sec:diss}

\paragraph{Scalability Advantage Compared to Standard NCDEs.}
In addition to the good empirical results shown above, our FWP approach also addresses an important limitation of existing NCDEs \citep{KidgerMFL20}: their
scalability in terms of model size.
The vector field in standard NCDEs (Eq.~\ref{eq:cde}) requires an NN $\mF_\theta$ which takes a vector $\vh(s) \in \mathbb{R}^d$ as an input to produce a matrix of size $\mathbb{R}^{d \times d_{\text{in}}}$.
This can be very challenging when $d_{\text{in}}$ or/and $d$ is large.
Actually, the same bottleneck is present in the weight generation of FWPs \citep{Schmidhuber:91fastweights}.
The use of outer products can remediate this issue in discrete FWPs as well as in CT FWPs:
the computations in our FWP-based NODE/NCDEs
only involve ``first-order'' dimensions (i.e., no multiplication between different dimensions, such as $d \times d_{\text{in}}$) for NN outputs.
This can scale well with increased model size, making feasible larger scale tasks  infeasible for existing NCDEs.
On the other hand, \citet{KidgerMFL20} report that using outer products (in their ``Sec.~6.1 on limitations'') in standard NCDEs does not perform well.
Why do outer products work well in our models but not in the original NCDEs?
The answer may be simple.
In the original NCDEs (Eq.~\ref{eq:cde}), multiplications occur at each (infinitesimal) time step between the generated rank-one weight matrix $\mF_\theta(\vh(s))$ and $\vx'(s)$ before the sum.
All these transformations are thus of rank one while we expect expressive transformations to be necessary to translate $\vx'(s)$ into changes in the state space.
In contrast, in our CT FWPs, the ODE only parameterises the weight generation process of another net, and thus the rank-one matrices are never used in isolation: they are summed up over time (Eq.~\ref{eq:ode-lr} or \ref{eq:cde-lr-x-dx}) to form an expressive weight matrix which is only then used for matrix multiplication (Eq. \ref{eq:querying_2} or \ref{eq:cde_fwp_query}.
The proposed FWP-NODE/NCDEs thus offer scalable alternatives to existing NCDEs, also yielding good empirical performance (Sec.~\ref{sec:exp}).

\paragraph{Importance of Memory Efficient Backpropagation for FWPs.}
Memory efficiency of continuous adjoint backpropagation 
may be not so important for standard NCDEs of state size $O(d)$, but is crucial for FWPs of state size  $O(d^2)$ which can quickly become prohibitive for long sequences, as naive backpropagation stores all states used in the forward pass.
Prior works on discrete FWPs \citep{schlag2021linear, irie2021training, irie2021going} solve this problem by a custom memory-efficient implementation. Here, the continuous adjoint method naturally addresses this problem.\looseness=-1

\paragraph{Limitations.}
Our ablation studies and hyper-parameter tuning
focus on optimising the model configuration/architecture.
From the NODE perspective, other parameters may further improve performance or alleviate performance variability/stability issues observed in some cases (Sec.~\ref{sec:exp}).
For example, we use the numerical solver configurations of the baselines (see Appendix \ref{app:exp_detail}) without tuning them.
Similarly, we use natural cubic splines to construct the differentiable control signals for NCDEs in the Speech Commands and PhysioNet Sepsis tasks, following the original NCDE paper \citep{KidgerMFL20}.
\citet{morrill2021neural} report performance enhancements by improving the corresponding interpolation methods (e.g., 93.7\% on Speech Commands).
Such further optimisation is not conducted here.

Generally speaking, real benefits of using continuous-time sequence processing models are yet to be proved.
While we achieve improvements over the best existing NODE/NCDE models on multiple datasets, discrete-time models tailored to the corresponding problem still perform as well or even better than our improved CT models (e.g., \citet{rusch2021long} for EigenWorms and \citet{che2018recurrent} for PhysioNet Sepsis; see Sec.~\ref{sec:exp}).

\paragraph{Related Work on Parameter/Weight ODEs.}
Other works  use ODEs to parameterise the time-evolving weights of some model.
However, they are limited to autonomous ODEs (i.e., no external control $\vx$ is involved).
\citet{ZhangYGGKMB19} and \citet{ChoromanskiDLSS20}
study coupled ODEs where one ODE is used for temporal evolution of parameters of the main Neural ODE.
The scope of these two works is limited to autonomous ODEs corresponding to continuous-depth residual NNs with different parameters per depth.
\citet{deleu2022continuous} consider an ODE version of a gradient descent learning process for adaptation, but also formulated as an autonomous ODE.
In contrast, our focus is really on \textit{sequence processing} where the model continuously receives external controls $\vx$ and translates them into weight changes of another network.

\section{Conclusion}
\label{sec:ccl}
We introduced novel continuous-time sequence processing neural networks that learn to use sequences of ODE-based continuous learning rules as elementary programming instructions to manipulate short-term memory in rapidly changing synaptic connections of another network.
The proposed models are continuous-time counterparts of Fast Weight Programmers and linear Transformers.
Our new models experimentally outperform by a large margin existing Neural ODE based sequence processors on very long or irregularly sampled time series.
Our Neural ODE/CDE based FWPs also address the fundamental scalability problem of the original Neural CDEs, which is highly promising for future applications of ODE based sequence processors to large scale problems.

\section{Acknowledgements}
We would like to thank \citet{KidgerMFL20}, \citet{morrillSKF21} and \citet{du2020model} for their public code.
This research was partially funded by ERC Advanced grant no: 742870, project AlgoRNN,
and by Swiss National Science Foundation grant no: 200021\_192356, project NEUSYM.
We are thankful for hardware donations from NVIDIA and IBM.
The resources used for this work were partially provided by Swiss National Supercomputing Centre (CSCS) project s1145 and s1154.

\bibliography{references}

\begin{thebibliography}{75}
\providecommand{\natexlab}[1]{#1}
\providecommand{\url}[1]{\texttt{#1}}
\expandafter\ifx\csname urlstyle\endcsname\relax
  \providecommand{\doi}[1]{doi: #1}\else
  \providecommand{\doi}{doi: \begingroup \urlstyle{rm}\Url}\fi

\bibitem[Chen et~al.(2018)Chen, Rubanova, Bettencourt, and Duvenaud]{ChenRBD18}
Tian~Qi Chen, Yulia Rubanova, Jesse Bettencourt, and David Duvenaud.
\newblock Neural ordinary differential equations.
\newblock In \emph{Proc. Advances in Neural Information Processing Systems
  (NeurIPS)}, pages 6572--6583, Montr{\'{e}}al, Canada, December 2018.

\bibitem[He et~al.(2016)He, Zhang, Ren, and Sun]{resnet}
Kaiming He, Xiangyu Zhang, Shaoqing Ren, and Jian Sun.
\newblock Deep residual learning for image recognition.
\newblock In \emph{{IEEE} Conf. on Computer Vision and Pattern Recognition
  ({CVPR})}, pages 770--778, Las Vegas, {NV}, {USA}, June 2016.

\bibitem[Srivastava et~al.(2015)Srivastava, Greff, and
  Schmidhuber]{srivastava2015icml}
Rupesh~K Srivastava, Klaus Greff, and J{\"u}rgen Schmidhuber.
\newblock Highway networks.
\newblock In \emph{the Deep Learning workshop at Int. Conf. on Machine Learning
  (ICML)}, Lille, France, July 2015.

\bibitem[Kidger et~al.(2020)Kidger, Morrill, Foster, and Lyons]{KidgerMFL20}
Patrick Kidger, James Morrill, James Foster, and Terry~J. Lyons.
\newblock Neural controlled differential equations for irregular time series.
\newblock In \emph{Proc. Advances in Neural Information Processing Systems
  (NeurIPS)}, Virtual Only, December 2020.

\bibitem[Elman(1990)]{elman1990finding}
Jeffrey~L Elman.
\newblock Finding structure in time.
\newblock \emph{Cognitive science}, 14\penalty0 (2):\penalty0 179--211, 1990.

\bibitem[Vaswani et~al.(2017)Vaswani, Shazeer, Parmar, Uszkoreit, Jones, Gomez,
  Kaiser, and Polosukhin]{trafo}
Ashish Vaswani, Noam Shazeer, Niki Parmar, Jakob Uszkoreit, Llion Jones,
  Aidan~N Gomez, {\L}ukasz Kaiser, and Illia Polosukhin.
\newblock Attention is all you need.
\newblock In \emph{Proc. Advances in Neural Information Processing Systems
  (NIPS)}, pages 5998--6008, Long Beach, {CA}, {USA}, December 2017.

\bibitem[Schmidhuber(1991)]{Schmidhuber:91fastweights}
J\"urgen Schmidhuber.
\newblock Learning to control fast-weight memories: An alternative to recurrent
  nets.
\newblock Technical Report FKI-147-91, Institut f\"{u}r Informatik, Technische
  Universit\"{a}t M\"{u}nchen, March 1991.

\bibitem[Schmidhuber(1992)]{schmidhuber1992learning}
J{\"u}rgen Schmidhuber.
\newblock Learning to control fast-weight memories: An alternative to dynamic
  recurrent networks.
\newblock \emph{Neural Computation}, 4\penalty0 (1):\penalty0 131--139, 1992.

\bibitem[Schlag et~al.(2021)Schlag, Irie, and Schmidhuber]{schlag2021linear}
Imanol Schlag, Kazuki Irie, and J\"urgen Schmidhuber.
\newblock Linear {T}ransformers are secretly fast weight programmers.
\newblock In \emph{Proc. Int. Conf. on Machine Learning (ICML)}, Virtual only,
  July 2021.

\bibitem[Irie et~al.(2021)Irie, Schlag, Csord\'as, and
  Schmidhuber]{irie2021going}
Kazuki Irie, Imanol Schlag, R\'obert Csord\'as, and J\"urgen Schmidhuber.
\newblock Going beyond linear transformers with recurrent fast weight
  programmers.
\newblock In \emph{Proc. Advances in Neural Information Processing Systems
  (NeurIPS)}, Virtual only, December 2021.

\bibitem[Katharopoulos et~al.(2020)Katharopoulos, Vyas, Pappas, and
  Fleuret]{katharopoulos2020transformers}
Angelos Katharopoulos, Apoorv Vyas, Nikolaos Pappas, and Fran{\c{c}}ois
  Fleuret.
\newblock Transformers are {RNN}s: Fast autoregressive transformers with linear
  attention.
\newblock In \emph{Proc. Int. Conf. on Machine Learning (ICML)}, Virtual only,
  July 2020.

\bibitem[Funahashi and Nakamura(1993)]{funahashi1993approximation}
Ken-ichi Funahashi and Yuichi Nakamura.
\newblock Approximation of dynamical systems by continuous time recurrent
  neural networks.
\newblock \emph{Neural networks}, 6\penalty0 (6):\penalty0 801--806, 1993.

\bibitem[Lapedes and Farber(1987)]{lapedes1987nonlinear}
Alan Lapedes and Robert Farber.
\newblock Nonlinear signal processing using neural networks: Prediction and
  system modelling.
\newblock \emph{Technical Report No. LA-UR-87-2662}, 1987.

\bibitem[Pineda(1987)]{pineda1987generalization}
Fernando~J Pineda.
\newblock Generalization of back-propagation to recurrent neural networks.
\newblock \emph{Physical review letters}, 59\penalty0 (19):\penalty0 2229,
  1987.

\bibitem[Pearlmutter(1989)]{pearlmutter1989learning}
Barak~A Pearlmutter.
\newblock Learning state space trajectories in recurrent neural networks.
\newblock \emph{Neural Computation}, 1\penalty0 (2):\penalty0 263--269, 1989.

\bibitem[Sato and Murakami(1991)]{sato1991learning}
Masa-aki Sato and Yoshihiko Murakami.
\newblock Learning nonlinear dynamics by recurrent neural.
\newblock In \emph{Some Problems on the Theory of Dynamical Systems in Applied
  Sciences-Proceedings of the Symposium}, volume~10, page~49, 1991.

\bibitem[Rico-Martinez et~al.(1992)Rico-Martinez, Krischer, Kevrekidis, Kube,
  and Hudson]{rico1992discrete}
Ramiro Rico-Martinez, K~Krischer, Ioannis Kevrekidis, MC~Kube, and JL~Hudson.
\newblock Discrete-vs. continuous-time nonlinear signal processing of {C}u
  electrodissolution data.
\newblock \emph{Chemical Engineering Communications}, 118\penalty0
  (1):\penalty0 25--48, 1992.

\bibitem[Oja(1982)]{oja1982simplified}
Erkki Oja.
\newblock Simplified neuron model as a principal component analyzer.
\newblock \emph{Journal of mathematical biology}, 15\penalty0 (3):\penalty0
  267--273, 1982.

\bibitem[Pearson(1901)]{pearson1901liii}
Karl Pearson.
\newblock {LIII}. {O}n lines and planes of closest fit to systems of points in
  space.
\newblock \emph{The London, Edinburgh, and Dublin philosophical magazine and
  journal of science}, 2\penalty0 (11):\penalty0 559--572, 1901.

\bibitem[Hotelling(1933)]{hotelling1933analysis}
Harold Hotelling.
\newblock Analysis of a complex of statistical variables into principal
  components.
\newblock \emph{Journal of educational psychology}, 24\penalty0 (6):\penalty0
  417, 1933.

\bibitem[Oja and Karhunen(1985)]{oja1985stochastic}
Erkki Oja and Juha Karhunen.
\newblock On stochastic approximation of the eigenvectors and eigenvalues of
  the expectation of a random matrix.
\newblock \emph{Journal of mathematical analysis and applications},
  106\penalty0 (1):\penalty0 69--84, 1985.

\bibitem[Oja(1989)]{oja1989neural}
Erkki Oja.
\newblock Neural networks, principal components, and subspaces.
\newblock \emph{International journal of neural systems}, 1\penalty0
  (01):\penalty0 61--68, 1989.

\bibitem[Plumbley(1995)]{plumbley1995lyapunov}
Mark~D Plumbley.
\newblock Lyapunov functions for convergence of principal component algorithms.
\newblock \emph{Neural Networks}, 8\penalty0 (1):\penalty0 11--23, 1995.

\bibitem[Hornik and Kuan(1992)]{hornik1992convergence}
Kurt Hornik and C-M Kuan.
\newblock Convergence analysis of local feature extraction algorithms.
\newblock \emph{Neural Networks}, 5\penalty0 (2):\penalty0 229--240, 1992.

\bibitem[Sanger(1989)]{sanger1989optimal}
Terence~D Sanger.
\newblock Optimal unsupervised learning in a single-layer linear feedforward
  neural network.
\newblock \emph{Neural networks}, 2\penalty0 (6):\penalty0 459--473, 1989.

\bibitem[{Wyatt Jr.} and Elfadel(1995)]{WyattE95}
John~L. {Wyatt Jr.} and Ibrahim~M. Elfadel.
\newblock Time-domain solutions of {O}ja's equations.
\newblock \emph{Neural Computation}, 7\penalty0 (5):\penalty0 915--922, 1995.

\bibitem[Fort and Pages(1996)]{fort1996convergence}
Jean-Claude Fort and Gilles Pages.
\newblock Convergence of stochastic algorithms: From the {K}ushner--{C}lark
  theorem to the {L}yapounov functional method.
\newblock \emph{Advances in applied probability}, 28\penalty0 (4):\penalty0
  1072--1094, 1996.

\bibitem[E(2017)]{weinan2017proposal}
Weinan E.
\newblock A proposal on machine learning via dynamical systems.
\newblock \emph{Communications in Mathematics and Statistics}, 1\penalty0
  (5):\penalty0 1--11, March 2017.

\bibitem[Haber and Ruthotto(2017)]{Haber_2017}
Eldad Haber and Lars Ruthotto.
\newblock Stable architectures for deep neural networks.
\newblock \emph{Inverse Problems}, 34\penalty0 (1):\penalty0 014004, December
  2017.

\bibitem[Haber et~al.(2018)Haber, Ruthotto, Holtham, and Jun]{HaberRHJ18}
Eldad Haber, Lars Ruthotto, Elliot Holtham, and Seong{-}Hwan Jun.
\newblock Learning across scales - multiscale methods for convolution neural
  networks.
\newblock In \emph{Proc. {AAAI} Conf. on Artificial Intelligence}, pages
  3142--3148, New Orleans, {LA}, {USA}, February 2018.

\bibitem[Chang et~al.(2018{\natexlab{a}})Chang, Meng, Haber, Ruthotto, Begert,
  and Holtham]{ChangMHRBH18}
Bo~Chang, Lili Meng, Eldad Haber, Lars Ruthotto, David Begert, and Elliot
  Holtham.
\newblock Reversible architectures for arbitrarily deep residual neural
  networks.
\newblock In \emph{Proc. {AAAI} Conf. on Artificial Intelligence}, pages
  2811--2818, New Orleans, {LA}, {USA}, February 2018{\natexlab{a}}.

\bibitem[Lu et~al.(2018)Lu, Zhong, Li, and Dong]{LuZLD18}
Yiping Lu, Aoxiao Zhong, Quanzheng Li, and Bin Dong.
\newblock Beyond finite layer neural networks: Bridging deep architectures and
  numerical differential equations.
\newblock In \emph{Proc. Int. Conf. on Machine Learning (ICML)}, pages
  3282--3291, Stockholm, Sweden, July 2018.

\bibitem[Chang et~al.(2018{\natexlab{b}})Chang, Meng, Haber, Tung, and
  Begert]{ChangMHTB18}
Bo~Chang, Lili Meng, Eldad Haber, Frederick Tung, and David Begert.
\newblock Multi-level residual networks from dynamical systems view.
\newblock In \emph{Int. Conf. on Learning Representations (ICLR)}, Vancouver,
  Canada, April 2018{\natexlab{b}}.

\bibitem[Ciccone et~al.(2018)Ciccone, Gallieri, Masci, Osendorfer, and
  Gomez]{CicconeGMOG18}
Marco Ciccone, Marco Gallieri, Jonathan Masci, Christian Osendorfer, and
  Faustino~J. Gomez.
\newblock {NAIS}-{N}et: Stable deep networks from non-autonomous differential
  equations.
\newblock In \emph{Proc. Advances in Neural Information Processing Systems
  (NeurIPS)}, pages 3029--3039, Montr{\'{e}}al, Canada, December 2018.

\bibitem[Pontryagin et~al.(1962)Pontryagin, Boltyanskii, Gamkrelidze, and
  Mishchenko]{pontryagin1962}
Liev~S. Pontryagin, VG~Boltyanskii, RV~Gamkrelidze, and EF~Mishchenko.
\newblock \emph{LS Pontryagin Selected Works: The Mathematical Theory of
  Optimal Processes}.
\newblock 1962.

\bibitem[Morrill et~al.(2021{\natexlab{a}})Morrill, Kidger, Yang, and
  Lyons]{morrill2021neural}
James Morrill, Patrick Kidger, Lingyi Yang, and Terry Lyons.
\newblock Neural controlled differential equations for online prediction tasks.
\newblock \emph{Preprint arXiv:2106.11028}, 2021{\natexlab{a}}.

\bibitem[Morrill et~al.(2021{\natexlab{b}})Morrill, Salvi, Kidger, and
  Foster]{morrillSKF21}
James Morrill, Cristopher Salvi, Patrick Kidger, and James Foster.
\newblock Neural rough differential equations for long time series.
\newblock In \emph{Proc. Int. Conf. on Machine Learning (ICML)}, pages
  7829--7838, Virtual only, July 2021{\natexlab{b}}.

\bibitem[Kidger(2021)]{kidger2022neural}
Patrick Kidger.
\newblock \emph{On Neural Differential Equations}.
\newblock PhD thesis, Mathematical Institute, University of Oxford, 2021.

\bibitem[Schmidhuber(1993)]{schmidhuber1993reducing}
J{\"u}rgen Schmidhuber.
\newblock Reducing the ratio between learning complexity and number of time
  varying variables in fully recurrent nets.
\newblock In \emph{International Conference on Artificial Neural Networks
  (ICANN)}, pages 460--463, Amsterdam, Netherlands, September 1993.

\bibitem[Ha et~al.(2017)Ha, Dai, and Le]{ha2017hypernetworks}
David Ha, Andrew Dai, and Quoc~V Le.
\newblock Hypernetworks.
\newblock In \emph{Int. Conf. on Learning Representations (ICLR)}, Toulon,
  France, April 2017.

\bibitem[Rubanova et~al.(2019)Rubanova, Chen, and Duvenaud]{RubanovaCD19}
Yulia Rubanova, Tian~Qi Chen, and David Duvenaud.
\newblock Latent ordinary differential equations for irregularly-sampled time
  series.
\newblock In \emph{Proc. Advances in Neural Information Processing Systems
  (NeurIPS)}, pages 5321--5331, Vancouver, Canada, December 2019.

\bibitem[Brouwer et~al.(2019)Brouwer, Simm, Arany, and Moreau]{BrouwerSAM19}
Edward~De Brouwer, Jaak Simm, Adam Arany, and Yves Moreau.
\newblock {GRU}-{ODE}-{B}ayes: Continuous modeling of sporadically-observed
  time series.
\newblock In \emph{Proc. Advances in Neural Information Processing Systems
  (NeurIPS)}, pages 7377--7388, Vancouver, Canada, December 2019.

\bibitem[Hebb(1949)]{hebb1949organization}
Donald~Olding Hebb.
\newblock The organization of behavior; a neuropsycholocigal theory.
\newblock \emph{A Wiley Book in Clinical Psychology}, 62:\penalty0 78, 1949.

\bibitem[Chou and Wang(2020)]{ChouW20}
Chi{-}Ning Chou and Mien~Brabeeba Wang.
\newblock {ODE}-inspired analysis for the biological version of {O}ja's rule in
  solving streaming {PCA}.
\newblock In \emph{Proc. Conf. on Learning Theory (COLT)}, pages 1339--1343,
  Virtual only, July 2020.

\bibitem[Zhang et~al.(2014)Zhang, Wang, and Liu]{zhang2014comprehensive}
Huaguang Zhang, Zhanshan Wang, and Derong Liu.
\newblock A comprehensive review of stability analysis of continuous-time
  recurrent neural networks.
\newblock \emph{IEEE Transactions on Neural Networks and Learning Systems},
  25\penalty0 (7):\penalty0 1229--1262, 2014.

\bibitem[Fermanian et~al.(2021)Fermanian, Marion, Vert, and
  Biau]{fermanian2021framing}
Adeline Fermanian, Pierre Marion, Jean-Philippe Vert, and G{\'e}rard Biau.
\newblock Framing {RNN} as a kernel method: A neural {ODE} approach.
\newblock In \emph{Proc. Advances in Neural Information Processing Systems
  (NeurIPS)}, Virtual only, December 2021.

\bibitem[Heusel et~al.(2017)Heusel, Ramsauer, Unterthiner, Nessler, and
  Hochreiter]{HeuselRUNH17}
Martin Heusel, Hubert Ramsauer, Thomas Unterthiner, Bernhard Nessler, and Sepp
  Hochreiter.
\newblock Gans trained by a two time-scale update rule converge to a local nash
  equilibrium.
\newblock In \emph{Proc. Advances in Neural Information Processing Systems
  (NIPS)}, pages 6626--6637, Long Beach, CA, USA, December 2017.

\bibitem[Aizerman et~al.(1964)Aizerman, Braverman, and
  Rozonoer]{aizerman1964theoretical}
Mark~A. Aizerman, Emmanuil~M. Braverman, and Lev~I. Rozonoer.
\newblock Theoretical foundations of potential function method in pattern
  recognition.
\newblock \emph{Automation and Remote Control}, 25\penalty0 (6):\penalty0
  917--936, 1964.

\bibitem[Irie et~al.(2022)Irie, Csord{\'a}s, and Schmidhuber]{irie2022dual}
Kazuki Irie, R{\'o}bert Csord{\'a}s, and J{\"u}rgen Schmidhuber.
\newblock The dual form of neural networks revisited: Connecting test time
  predictions to training patterns via spotlights of attention.
\newblock In \emph{Proc. Int. Conf. on Machine Learning (ICML)}, Baltimore,
  {MD}, {USA}, July 2022.

\bibitem[Widrow and Hoff(1960)]{widrow1960adaptive}
Bernard Widrow and Marcian~E Hoff.
\newblock Adaptive switching circuits.
\newblock In \emph{Proc. {IRE} WESCON Convention Record}, pages 96--104, Los
  Angeles, {CA}, {USA}, August 1960.

\bibitem[Warden(2018)]{warden2018speech}
Pete Warden.
\newblock Speech commands: A dataset for limited-vocabulary speech recognition.
\newblock \emph{Preprint arXiv:1804.03209}, 2018.

\bibitem[Reyna et~al.(2019)Reyna, Josef, Seyedi, Jeter, Shashikumar, Westover,
  Sharma, Nemati, and Clifford]{ReynaJSJSWSNC19}
Matthew~A. Reyna, Christopher Josef, Salman Seyedi, Russell Jeter, Supreeth~P.
  Shashikumar, M.~Brandon Westover, Ashish Sharma, Shamim Nemati, and Gari~D.
  Clifford.
\newblock Early prediction of sepsis from clinical data: the
  {P}hysio{N}et/computing in cardiology challenge 2019.
\newblock In \emph{Proc. Computing in Cardiology (CinC)}, pages 1--4,
  Singapore, September 2019.

\bibitem[Bagnall et~al.(2018)Bagnall, Dau, Lines, Flynn, Large, Bostrom,
  Southam, and Keogh]{bagnall2018uea}
Anthony Bagnall, Hoang~Anh Dau, Jason Lines, Michael Flynn, James Large, Aaron
  Bostrom, Paul Southam, and Eamonn Keogh.
\newblock The {UEA} multivariate time series classification archive, 2018.
\newblock \emph{Preprint arXiv:1811.00075}, 2018.

\bibitem[Che et~al.(2018)Che, Purushotham, Cho, Sontag, and
  Liu]{che2018recurrent}
Zhengping Che, Sanjay Purushotham, Kyunghyun Cho, David Sontag, and Yan Liu.
\newblock Recurrent neural networks for multivariate time series with missing
  values.
\newblock \emph{Scientific reports}, 8\penalty0 (1):\penalty0 1--12, 2018.

\bibitem[Rusch and Mishra(2021)]{rusch2021unicornn}
T~Konstantin Rusch and Siddhartha Mishra.
\newblock Un{ICORNN}: A recurrent model for learning very long time
  dependencies.
\newblock In \emph{Proc. Int. Conf. on Machine Learning (ICML)}, pages
  9168--9178, Virtual only, July 2021.

\bibitem[Rusch et~al.(2022)Rusch, Mishra, Erichson, and Mahoney]{rusch2021long}
T~Konstantin Rusch, Siddhartha Mishra, N~Benjamin Erichson, and Michael~W
  Mahoney.
\newblock Long expressive memory for sequence modeling.
\newblock In \emph{Int. Conf. on Learning Representations (ICLR)}, Virtual
  only, April 2022.

\bibitem[Irie and Schmidhuber(2021)]{irie2021training}
Kazuki Irie and J{\"u}rgen Schmidhuber.
\newblock Training and generating neural networks in compressed weight space.
\newblock In \emph{ICLR Neural Compression Workshop}, Virtual only, May 2021.

\bibitem[Zhang et~al.(2019)Zhang, Yao, Gholami, Gonzalez, Keutzer, Mahoney, and
  Biros]{ZhangYGGKMB19}
Tianjun Zhang, Zhewei Yao, Amir Gholami, Joseph~E. Gonzalez, Kurt Keutzer,
  Michael~W. Mahoney, and George Biros.
\newblock {ANODEV2:} {A} coupled neural {ODE} framework.
\newblock In \emph{Proc. Advances in Neural Information Processing Systems
  (NeurIPS)}, pages 5152--5162, Vancouver, Canada, December 2019.

\bibitem[Choromanski et~al.(2020)Choromanski, Davis, Likhosherstov, Song,
  Slotine, Varley, Lee, Weller, and Sindhwani]{ChoromanskiDLSS20}
Krzysztof~Marcin Choromanski, Jared~Quincy Davis, Valerii Likhosherstov,
  Xingyou Song, Jean{-}Jacques~E. Slotine, Jacob Varley, Honglak Lee, Adrian
  Weller, and Vikas Sindhwani.
\newblock Ode to an {ODE}.
\newblock In \emph{Proc. Advances in Neural Information Processing Systems
  (NeurIPS)}, Virtual only, December 2020.

\bibitem[Deleu et~al.(2022)Deleu, Kanaa, Feng, Kerg, Bengio, Lajoie, and
  Bacon]{deleu2022continuous}
Tristan Deleu, David Kanaa, Leo Feng, Giancarlo Kerg, Yoshua Bengio, Guillaume
  Lajoie, and Pierre-Luc Bacon.
\newblock Continuous-time meta-learning with forward mode differentiation.
\newblock In \emph{Int. Conf. on Learning Representations (ICLR)}, Virtual
  only, April 2022.

\bibitem[Du et~al.(2020)Du, Futoma, and Doshi-Velez]{du2020model}
Jianzhun Du, Joseph Futoma, and Finale Doshi-Velez.
\newblock Model-based reinforcement learning for semi-{M}arkov decision
  processes with neural {ODE}s.
\newblock In \emph{Proc. Advances in Neural Information Processing Systems
  (NeurIPS)}, Virtual only, 2020.

\bibitem[Ba et~al.(2016)Ba, Kiros, and Hinton]{ba2016layer}
Jimmy~Lei Ba, Jamie~Ryan Kiros, and Geoffrey~E Hinton.
\newblock Layer normalization.
\newblock \emph{Preprint arXiv:1607.06450}, 2016.

\bibitem[Irie et~al.(2019)Irie, Zeyer, Schl\"uter, and Ney]{irie19:trafolm}
Kazuki Irie, Albert Zeyer, Ralf Schl\"uter, and Hermann Ney.
\newblock Language modeling with deep {T}ransformers.
\newblock In \emph{Proc. Interspeech}, pages 3905--3909, Graz, Austria,
  September 2019.

\bibitem[Todorov et~al.(2012)Todorov, Erez, and Tassa]{todorovET12}
Emanuel Todorov, Tom Erez, and Yuval Tassa.
\newblock Mu{J}o{C}o: {A} physics engine for model-based control.
\newblock In \emph{{IEEE/RSJ} Int. Conf. on Intelligent Robots and Systems
  ({IROS})}, pages 5026--5033, Vilamoura, Portugal, October 2012. {IEEE}.

\bibitem[Sharma et~al.(2017)Sharma, Lakshminarayanan, and
  Ravindran]{SharmaLR17}
Sahil Sharma, Aravind~S. Lakshminarayanan, and Balaraman Ravindran.
\newblock Learning to repeat: Fine grained action repetition for deep
  reinforcement learning.
\newblock In \emph{Int. Conf. on Learning Representations (ICLR)}, Toulon,
  France,, April 2017.

\bibitem[Puterman(1994)]{puterman2014markov}
Martin~L Puterman.
\newblock \emph{Markov decision processes: discrete stochastic dynamic
  programming}.
\newblock 1994.

\bibitem[Richards(2005)]{richards2005robust}
Arthur~George Richards.
\newblock \emph{Robust constrained model predictive control}.
\newblock PhD thesis, Massachusetts Institute of Technology, 2005.

\bibitem[Camacho and Bordons(2013)]{camacho2013model}
Eduardo~F Camacho and Carlos Bordons.
\newblock \emph{Model predictive control}.
\newblock 2013.

\bibitem[Mayne et~al.(2000)Mayne, Rawlings, Rao, and
  Scokaert]{mayne2000constrained}
David~Q Mayne, James~B Rawlings, Christopher~V Rao, and Pierre~OM Scokaert.
\newblock Constrained model predictive control: Stability and optimality.
\newblock \emph{Automatica}, 36\penalty0 (6):\penalty0 789--814, 2000.

\bibitem[Lillicrap et~al.(2016)Lillicrap, Hunt, Pritzel, Heess, Erez, Tassa,
  Silver, and Wierstra]{lillicrapHPHETS15}
Timothy~P. Lillicrap, Jonathan~J. Hunt, Alexander Pritzel, Nicolas Heess, Tom
  Erez, Yuval Tassa, David Silver, and Daan Wierstra.
\newblock Continuous control with deep reinforcement learning.
\newblock In \emph{Int. Conf. on Learning Representations (ICLR)}, San Juan,
  Puerto Rico, May 2016.

\bibitem[Silver et~al.(2014)Silver, Lever, Heess, Degris, Wierstra, and
  Riedmiller]{Silver2014}
David Silver, Guy Lever, Nicolas Heess, Thomas Degris, Daan Wierstra, and
  Martin Riedmiller.
\newblock Deterministic policy gradient algorithms.
\newblock In \emph{Proc. Int. Conf. on Machine Learning (ICML)}, Beijing,
  China, June 2014.

\bibitem[Sutton(1984)]{sutton1984temporal}
Richard~S Sutton.
\newblock \emph{Temporal Credit Assignment in Reinforcement Learning}.
\newblock PhD thesis, University of Massachusetts Amherst, 1984.

\bibitem[Konda and Tsitsiklis(1999)]{kondaactorcritic}
Vijay~R. Konda and John~N. Tsitsiklis.
\newblock Actor-critic algorithms.
\newblock In \emph{Proc. Advances in Neural Information Processing Systems
  (NIPS)}, pages 1008--1014, Denver, CO, USA, November 1999.

\bibitem[Peters et~al.(2005)Peters, Vijayakumar, and Schaal]{PetersVS05}
Jan Peters, Sethu Vijayakumar, and Stefan Schaal.
\newblock Natural actor-critic.
\newblock In \emph{Proc. European Conference on Machine Learning (ECML)}, pages
  280--291, Porto, Portugal, October 2005.

\bibitem[Yildiz et~al.(2021)Yildiz, Heinonen, and
  L{\"{a}}hdesm{\"{a}}ki]{YildizHL21}
{\c{C}}agatay Yildiz, Markus Heinonen, and Harri L{\"{a}}hdesm{\"{a}}ki.
\newblock Continuous-time model-based reinforcement learning.
\newblock In \emph{Proc. Int. Conf. on Machine Learning (ICML)}, Virtual only,
  July 2021.

\end{thebibliography}
\bibliographystyle{unsrtnat}

\clearpage
\appendix

\section{Further Model Specifications}
\label{app:model}

For better readability,
Sec.~\ref{sec:ct_fwp} omitted element-wise activation functions for query/key/value vectors.
For example, for Eq.~\ref{eq:direct_ode}, the key and value vectors, $\vk$ and $\vv$, are generated as follows
\begin{align} 
\label{eq:direct_ode_act}
\beta(s), \vx_{\vk}(s), \vx_{\vv}(s) &= \mW_\text{slow} \vx(s) \\
\vk(s) &= \softmax(\vx_{\vk}(s)) \\
\label{eq:tanh_value}
\vv(s) &= \tanh(\vx_{\vv}(s))
\end{align}
Eq.~\ref{eq:querying_1} for the query generation has to be replaced by
\begin{align} 
\label{eq:direct_ode_act_query}
\vq(T) &= \softmax(\mW_q \vx(T))
\end{align}
These specifications are analogous in the CDE cases, i.e., in Eqs.~\ref{eq:cde_lr}-\ref{eq:cde_fwp_query}.

While  $\softmax$ has already been identified as a crucial component for stability in prior works on discrete-time FWPs \citep{schlag2021linear},
we  apply an additional $\tanh$ to the value vectors (Eq.~\ref{eq:tanh_value}).
Such usage of $\tanh$ in the output of vector fields for CDEs has been advocated by \citet{KidgerMFL20} to improve the stability of the ODE solver.
Indeed, we also generally found this beneficial, and sometimes crucial, for stable training of our models.
For the Delta rule (Eq.~\ref{eq:cde_lr}), since we modify the value vector after projection (to take into account the ``value'' which is currently associated with the key vector), we consider two different ways of applying $\tanh$ as follows:
\begin{align} 
\label{eq:cde_lr_delta_tanh}
\mF_\theta(\mW(s), \vx(s))
    &= \sigma(\beta(s))
    \begin{cases}
      \big(\tanh(\vx_{\vv}(s)) - \mW(s)\vk(s)\big) \otimes \vk(s)& \text{pre-delta} \\
      \tanh(\vx_{\vv}(s) - \mW(s)\vk(s)) \otimes \vk(s)& \text{post-delta}
    \end{cases}
\end{align}
For the task involving very long sequences ($>$ 4000 frames), we found the ``post-delta'' version to be crucial for successful training.

\section{Further Experimental Details}
\label{app:exp_detail}

Our basic settings for data preparation and training are based on those used by prior works \citep{KidgerMFL20, morrillSKF21} (we use their public implementations) for fair comparisons with the corresponding baselines.

\paragraph{Dataset/Task details.}
The essential information about the datasets has  already been presented in Sec.~\ref{sec:exp}.
The number of sequences in the training set is about 24\,K for the Speech Commands dataset and about 28\,K for the PhysioNet Sepsis dataset.

\paragraph{Model/Training details.}
As mentioned in Sec.~\ref{sec:ct_fwp},
all our FWP models make use of multiple computational heads, and their NODE/NCDE layer is followed by the standard Transformer feedforward block \citep{trafo}.
Layer normalisation \citep{ba2016layer} is also applied inside their vector field and feedforward block.
The number of heads $n_\text{head}$ and the feedforward inner dimension $d_\text{ff}$ are hyper-parameters of the model,
in addition to the size $d_\text{model}$ of the NODE/NCDE layer.
For all models considered in the 
main text, the number of layers is one (see Sec.~\ref{app:deeper} below for a discussion of deeper models).
We conducted hyper-parameter search in the following ranges, and selected the best configuration for each setting based on its validation performance.

For Speech Commands: $d_\text{model} \in \{32, 80, 128, 160\}$, $n_\text{head} \in \{4, 8, 16, 32\}$ and $d_\text{ff}=\{80, 4 * d_\text{model}\}$, with a learning rate $\eta \in \{1e-5, 5e-5\}$ and a batch size of 1024.
The best Delta CDE model is obtained for ($d_\text{model}=80$, $n_\text{head}=4$, $d_\text{ff}=320$, $\eta=5e-5$).

For PhysioNet Sepsis: $d_\text{model} \in \{32, 80, 128, 160\}$, $n_\text{head} \in \{8, 16, 32\}$ and $d_\text{ff} \in \{64, 4 * d_\text{model}\}$, with a learning rate $\eta \in \{1e-5, 2e-5, 3e-5, 4e-5, 5e-5, 6e-5, 8e-5, 1e-4\}$ and a batch size of 1024.
The best Delta CDE model is obtained for ($d_\text{model}=80$, $n_\text{head}=16$, $d_\text{ff}=64$, $\eta=3e-5$) in the ``OI'' case,
and for ($d_\text{model}=160$, $n_\text{head}=32$, $d_\text{ff}=64$, $\eta=4e-5$) in the ``no-OI'' setting.

For EigenWorms: $d_\text{model} \in \{64, 128, 160\}$, $n_\text{head} \in \{8, 16, 32\}$ and $d_\text{ff} \in \{32, 64, 80, 128\}$, with a learning rate $\eta \in \{5e-5, 1e-4, 3e-4, 5e-4\}$ and a batch size set to 181 to include all training sequences (70\% of the 259 total sequences in this dataset, again following the prior work \citep{morrillSKF21}).
The best overall Delta CDE model is obtained for ($d_\text{model}=128$, $n_\text{head}=16$, $d_\text{ff}=64$, $\eta=1e-4$) in the log-signature depth-1 case.

The best hyper-parameters for each model are listed in Table \ref{tab:hyper}.

\begin{table}[t]
\caption{Hyper-parameters. Their definitions can be found in Appendix \ref{app:exp_detail}. H/O/D denote Hebb/Oja/Delta variants respectively. The top part shows hyper-parameters of the direct NODE models for Speech Commands and PhysioNet Sepsis, and those of the NRDE models for EigenWorms. The bottom part is for the NCDE models.
}
\label{tab:hyper}
\setlength{\tabcolsep}{0.45em}
\begin{center}
\begin{tabular}{llrrrrrrrrrrrr}
\toprule
& & \multicolumn{3}{l}{\multirow{2}{*}{Speech Commands}} & \multicolumn{3}{l}{PhysioNet Sepsis} & \multicolumn{3}{l}{PhysioNet Sepsis} & \multicolumn{3}{l}{\multirow{2}{*}{EigenWorms}} \\ 
 & & & & & \multicolumn{3}{l}{OI} & \multicolumn{3}{l}{no-OI} & && \\ \midrule
Type & Param. & H & O & D & H & O & D & H & O & D & H & O & D \\ \midrule
O/RDE & $d_\text{model}$ & 128 & 128 & 160 & 160 & 160 & 128 & 32 & 160 & 80 & 160 & 160 & 128 \\
& $n_\text{head}$ & 32 & 32 & 16 & 32 & 32 & 16 & 8 & 32 & 16 & 16 & 16 &  16 \\
& $d_\text{ff}$   & 512 & 512 & 80 & 64 & 64  & 64 & 64 & 64 & 64  & 128 & 128 &  64 \\
& $\eta$ (1e-5)  & 5 & 5 & 1 & 6 & 6& 2 & 10 & 4 & 4 & 30 & 30 & 10 \\
\midrule \midrule
CDE & $d_\text{model}$ & 128 & 128 & 80 & 32 & 80 & 80 & 160 & 160 & 160 & 160 & 128 & 128 \\
& $n_\text{head}$ & 32 & 32 & 4 & 8 & 16 & 16 & 32 & 32 & 32 & 16 & 16 & 16 \\
& $d_\text{ff}$ & 512 & 512 & 320 & 256 & 64 & 64 & 64 & 64 & 64 & 128 & 64 & 64 \\
& $\eta$ (1e-5) & 5 & 5 & 5 & 6 & 2 & 3 & 4 & 6  & 4 & 30 & 30 & 10 \\ 
\bottomrule
\end{tabular}
\end{center}
\end{table}

\paragraph{Numerical solver details.}
For all tasks, we use the numerical solver settings of the baseline models.
For all supervised learning experiments in the main text, we use the `\texttt{rk4}' Runge-Kutta variant with default tolerance parameters in \url{https://github.com/rtqichen/torchdiffeq} \citep{ChenRBD18}: relative/absolute tolerance values are 1e-7/1e-9.
For the RL experiment (Sec.~\ref{app:rl}), `\texttt{dopri5}' is used instead, with relative/absolute tolerance values of 1e-5/1e-6.
As mentioned in Sec.~\ref{sec:diss},
we do not tune these configurations specifically for our models.
Further enhancements in terms of performance or stability may be obtained by tuning them.

For any further details, we refer to our code.

\section{Extra Experiments}
\label{app:exp}

\subsection{Deeper Models}
\label{app:deeper}
As the datasets used here were rather small,
one-layer models yield satisfactory performance.
In general, however, deeper architectures are common for Transformers \citep{irie19:trafolm}.
Here we show methods to increase the depth of FWP-based NODE/NCDE models.
While the possibility of stacking multiple NCDE layers has been mentioned previously \citep{kidger2022neural},
no practical result has been reported.\looseness=-1

The following equations are for the direct NODE case of our FWP models (the CDE case is analogous).
Let $L$ and $l$ denote positive integers.
The forward computation of layer $l$ in an $L$-layer model at step $t$ is as follows
\begin{align}
\mW(t, l) &= \mW(t_0, l) + \int_{s=t_0}^t \mF_{\theta_l}^{(l)}(\mW(s, l), \vx(s, l-1))ds \\
\label{eq:coupled_query}
\vx(t, l) &= \FFN\left(\mW(t, l)\softmax(\mW_q \vx(t, l-1))\right)
\end{align}
where $\FFN$ denotes the standard Transformer feedforward block, $\mW(t, l)$ denotes the fast weight matrix, $\mF_{\theta_l}^{(l)}$ is the vector field with parameters $\theta_l$, and $\vx(t, l)$ denotes the output of layer $l \geq 1$ while $\vx(t, 0)$ is the external control signal.
The output of layer $l$, $\vx(t, l)$, thus becomes the control signal for the next layer $l+1$, and so on.
This yields a system of $L$ \textit{coupled ODEs}.
The fast weight matrix for each layer at time $T$ can be obtained by calling the ODE solver for the corresponding coupled ODEs
\begin{align}
\mW(T, 1), ..., \mW(T, L) = \ODESolve([\mF_{\theta_1}^{1}, ..., \mF_{\theta_L}^{L}], [\mW(t_0, 1), ..., \mW(t_0, L)], t_0, T)
\end{align}

The final output $\vx(T, L)$ can be computed recursively via Eq.~\ref{eq:coupled_query} for $t=T$ starting from $l=1$.

We conduct experiments in the EigenWorms dataset with the goal of improving our best single-layer model of Table \ref{tab:eigenworms}.
We tune the 2-layer model using the same hyper-parameter search space used for the 1-layer model.
However, the best obtained 2-layer performance is 86.7 \% (4.1) which is worse than the best 1-layer performance of 91.8 \% (3.4).
A natural next step towards obtaining potential performance gains from deeper/larger models is to apply regularisation techniques to our models (we leave that to future work).

\subsection{Ablation Studies on CDE-based Models}
\label{app:cde_variants}
In Sec.~\ref{sec:cde-lr} on NCDE based FWPs, we mention two optional model variations skipped in the main text:
the Hebb variant which uses $\vx'$ to generate keys and $\vx$ to generate values, and NCDE based models which only use $\vx'$ (instead of $\vx$ and $\vx'$) in the vector field (see footnote \ref{foot:var}).
Here we provide the corresponding ablation studies.

\paragraph{Using $\vx'$ for key and $\vx$ for value generation in the Hebb variant.}
In Sec.~\ref{sec:cde-lr} on FWP based CDEs,
we note that there are two possible formulations for the Hebb variant.
The following equations highlight the difference between the two formulations (depending on the variable, $\vx$ or $\vx'$, used to generate keys/values):
\begin{align} 
\tF_\theta\big(\mW(s), \vx(s), \vx'(s)\big)\vx'(s)
    &= \sigma(\beta(s))
    \begin{cases}
     \mW_k \vx(s) \otimes \mW_v \vx'(s)  & \text{$\vx$-keys, $\vx'$-values as in Eq.~\ref{eq:cde_lr}}\\
     \mW_v \vx(s) \otimes \mW_k \vx'(s)  & \text{$\vx$-values, $\vx'$-keys}\\
    \end{cases}
\end{align}
We also use different equations for the fast net forward computation
where, for consistency, we generate the query from the same input ($\vx$ or $\vx'$) used to generate keys (even if this is not a strict requirement):

\begin{align} 
\vy(T) &=
    \begin{cases}
     \mW(T)^\top \mW_q \vx(T)& \text{$\vx$-keys, $\vx'$-values}\\
     \mW(T) \mW_q \vx'(T)  & \text{$\vx$-values, $\vx'$-keys}
    \end{cases}
\end{align}

Experimentally, we found this second variant (``$\vx$-values, $\vx'$-keys'') to perform worse on the Speech Commands task.
It obtained a test accuracy of 83.9 \% (0.4) compared to 89.5 \% (0.3) achieved by the first variant we reported in the main text (Table \ref{tab:speech}).

\paragraph{FWP based NCDEs using only $\vx'$.}
In Sec.~\ref{sec:cde-lr},
we note that tractable FWP based NCDEs can also be obtained by using only $\vx'$ in the vector field (instead of $\vx$ and $\vx'$).
The equations for the corresponding models can be obtained by replacing $\vx$ by $\vx'$ in Eq.~\ref{eq:cde_lr}:
\begin{align} 
\label{eq:cde_lr_dx_only}
\tF_\theta\big(\mW(s), \vx'(s)\big)\vx'(s)
    &= \sigma(\beta(s))
    \begin{cases}
     \mW_k \vx'(s) \otimes \mW_v \vx'(s)  & \text{Hebb}\\
      \big(\mW_k \vx'(s) - \mW(s)^\top \mW_v \vx'(s) \big) \otimes \mW_v \vx'(s) & \text{Oja} \\
      \big(\mW_v \vx'(s) - \mW(s)\mW_k \vx'(s)\big) \otimes \mW_k \vx'(s) & \text{Delta}
    \end{cases}
\end{align}

On the Speech Commands task, we found it crucial to use both $\vx$ and $\vx'$.
The second variants using only $\vx'$ perform much worse (< 65 \%) than the variants using both $\vx$ and $\vx'$, across all three learning rule schemes.

\subsection{Model-based Reinforcement Learning (RL)}
\label{app:rl}
The main focus of this paper is the one of \citet{KidgerMFL20} where it is assumed that at least some bounded (piece-wise) continuous control signal $\vx(t)$ defined on $[t_0, T]$ is available.
However, for the sake of completeness, in Sec.~\ref{sec:ode_rfwp}, we also present the ODE-RFWP and its latent variant in the case where we can only obtain discrete inputs.
Here we provide some experimental results with working examples of such models in a model-based RL  setting previously proposed by \citet{du2020model}.

\paragraph{Settings.}
We use the MuJoCo environments \citep{todorovET12}.
Our setting follows the version with action repetitions \citep{SharmaLR17}  proposed by \citet{du2020model},
formalised as semi-Markov decision processes \citep{puterman2014markov}. 
The core difference to the standard setting is that there is a (state-dependent) time gap $\tau > 0$ between the time when an agent takes an action in state $\vs$ and when it observes a new state $\vs'$ (and can take a new action), resulting in irregularly timed observations.
The time gap $\tau$ is a deterministic function of the state $\vs$.
\citet{du2020model} proposed a model-based planning approach where an environmental model is parameterised as an ODE-RNN or Latent ODE-RNN such that an ODE is used to model the latent state transitions between observations.
The model is trained to predict
state transitions (using mean squared error for ODE-RNNs and negative evidence lower bounds for Latent ODE-RNNs) and  time gaps between observations,
and it is used for model predictive control (MPC) \citep{richards2005robust, camacho2013model, mayne2000constrained}.
The agent is trained by Deep Deterministic Policy Gradient (DDPG) \citep{lillicrapHPHETS15, Silver2014} with actor-critic \citep{sutton1984temporal, kondaactorcritic, PetersVS05} where the predictions of policy and value networks are conditioned on the recurrent (latent) state of the environment model.
In \citet{du2020model}'s work, the Latent ODE-RNN was shown to outperform the ODE-RNN and the model-free baseline in terms of sample efficiency in \textit{Hopper},  \textit{Swimmer} and \textit{Half Cheetah} environments.
Therefore, we compare our Latent ODE-RFWP to the baseline Latent ODE-RNN.
We conducted our experiments in \textit{Hopper} and \textit{Swimmer} environments
(we excluded \textit{Half Cheetah}
which was reported \citep{du2020model} to require larger models and thus more compute).
On a side note, there are other works \citep{YildizHL21} with a focus on RL in continuous-time environments.
However, the available environments are still limited in terms of complexity (e.g., CartPole).

\paragraph{Training/Model details.}
Each training iteration consists of interactions with the environment to collect trajectories, training of the environmental model, and policy optimisation.
Following \citet{du2020model},
each iteration consists of 5\,K environmental steps,
and we train for up to 400\,K steps (corresponding to a total of 80 iterations) in addition to the initial random interactions of 15\,K steps.
The planning horizon for MPC is set to 10 steps.
We set the hidden state size of our RFWP models to 128 like in the baseline. We use 16 computational heads, and a feedforward block size of 512.
We found that we can use the same RNN-based encoder used in the baseline also for our RFWP models without notable difference in terms of performance
compared to an FWP-based encoder.
The same training/model configurations are used for both \textit{Swimmer} and \textit{Hopper}.
They are identical to those used by \citet{du2020model} (we use their base implementation), apart from the fact that we use the continuous adjoint method to train all models.
For further details, we refer to our code.

\begin{figure*}[ht]
	\subfloat[Swimmer]{
		\centering
		\includegraphics[width=.43\linewidth]{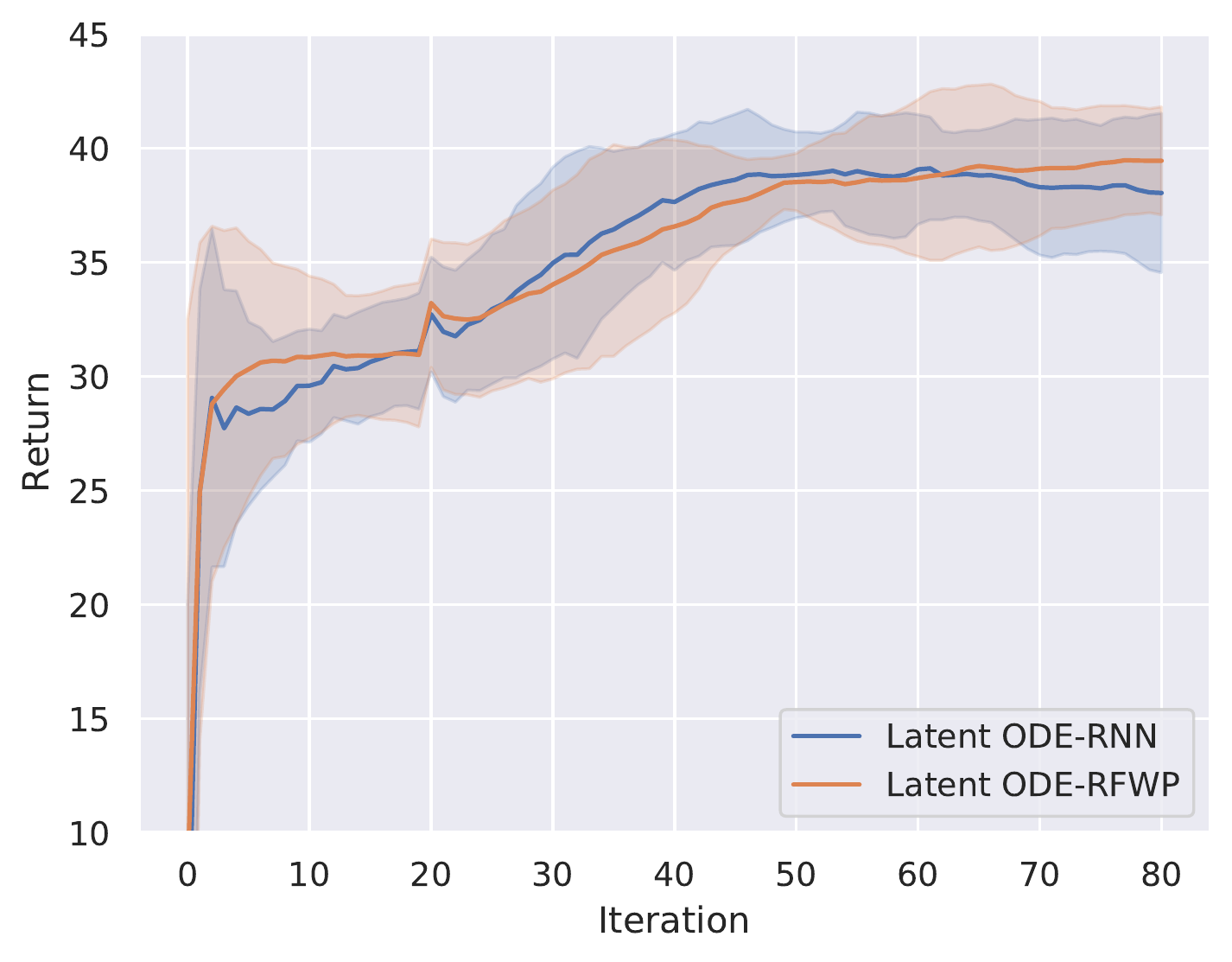}
		\label{sfig:swimmer}
	}
	\qquad
	\subfloat[Hopper]{
		\centering
		\vspace{-5mm}
		\includegraphics[width=.44\linewidth]{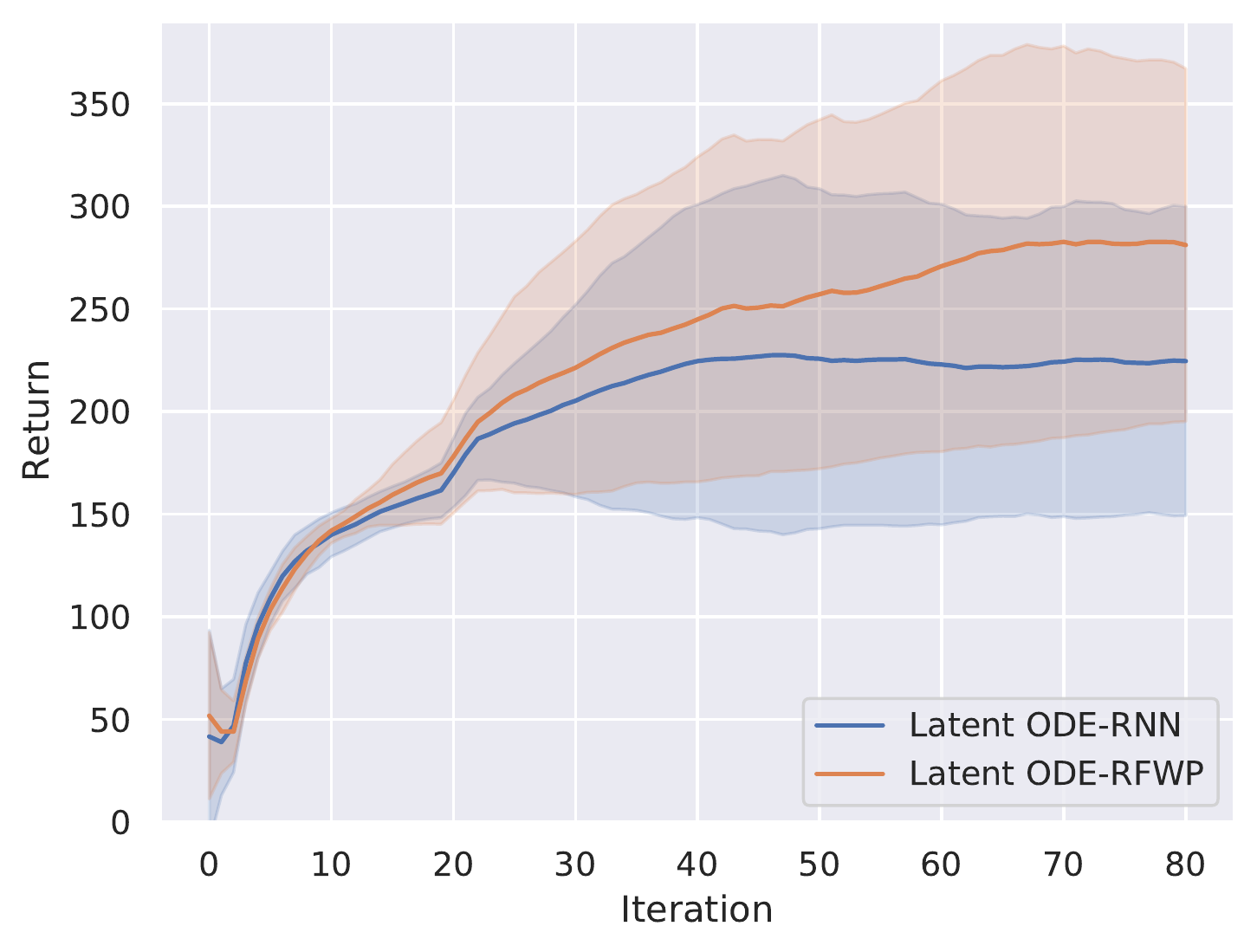}
		\label{sfig:hopper}
	}
        \caption{Return over training iterations in the \textbf{Swimmer} and \textbf{Hopper} environments, averaged over five test episodes. Smoothing is applied on overlapping windows of size 20. One iteration corresponds to  5\,K interactions with the environment. Mean/std are computed for ten runs.}
\end{figure*}

\paragraph{Results.}
Figures \ref{sfig:swimmer} and \ref{sfig:hopper} show learning curves of baseline Latent ODE-RNN and our RFWP variant in \textit{Swimmer} and \textit{Hopper} environments, respectively.
Given a high variability of results across seeds, we present an average over ten training runs.
We observe that final performance and sample efficiency of both models are comparable on \textit{Swimmer}, while the RFWP model yields better performance on \textit{Hopper}.
These results indicate that the Latent ODE-RFWP model presented in Sec.~\ref{sec:ode_rfwp} can effectively in practice be used as a replacement of the standard Latent ODE-RNN.
We note, however, that the final performance of baseline Latent ODE-RNN is below the one reported by \citet{du2020model} (which exceeds an expected return of 350).
The only change we introduced to the original setting is to consistently use the continuous adjoint method to train all models (further ablation studies may be needed to evaluate the impact of this change), and to use ten seeds instead of four.

\end{document}